%% file: iclr2027_conference.tex
\documentclass{article} 
\usepackage{iclr2027_conference,times}

\input{math_commands.tex}

\usepackage{xcolor}
\usepackage{titlesec}
\titlespacing*{\section}{0pt}{1.2ex}{0.8ex}
\usepackage{hyperref}
\usepackage{url}
\usepackage{booktabs}
\usepackage{amsmath}
\usepackage{algorithm} 
\usepackage{algpseudocode}
\usepackage{multirow}
\usepackage{booktabs}
\usepackage{graphicx}
\usepackage{amssymb}
\usepackage{adjustbox}
\usepackage{wrapfig}  
\usepackage{pgfplots}
\pgfplotsset{compat=1.18}

\usepackage[table]{xcolor}
\definecolor{mapit}{HTML}{f9e2ae}
\usepackage{enumitem}
\usepackage{titletoc}
\usepackage[bottom]{footmisc}
\usepackage{capt-of}
\usepackage[most]{tcolorbox}

\title{Text-to-Image Generative Modeling\\Without Language Pretraining}

\definecolor{linkcolor}{RGB}{0,102,204}

\author{Keshigeyan Chandrasegaran$^{1,2}$ \enspace Yue Zhao$^{1}$ \enspace Armin W.~Thomas$^{2}$ \\ \textbf{Ehsan Adeli}$^{1}$ \enspace \textbf{Jiajun Wu}$^{1}$ \enspace \textbf{Juan Carlos Niebles}$^{1}$ \enspace \textbf{Li Fei-Fei}$^{1}$ \enspace \textbf{Michael Poli}$^{1,2}$ \\
$^{1}$Stanford University \quad $^{2}$Radical Numerics
}

\definecolor{textcolor}{RGB}{40,90,160}
\definecolor{imagecolor}{RGB}{190,90,45}
\definecolor{specialcolor}{RGB}{90,90,90}

\newcommand{\spctok}[1]{\textcolor{specialcolor}{#1}}
\usepackage{array}

\definecolor{xgreen}{HTML}{1B9E77}    
\definecolor{xgreenbg}{HTML}{EEF7F3}  
\definecolor{darkgreen}{RGB}{0,158,115}
\usepackage{mapit_style}

\iclrfinalcopy 

\begin{document}

\maketitle
\begingroup
\renewcommand{\thefootnote}{}
\footnotetext{Correspondence to \texttt{keshik@cs.stanford.edu}. Code and models are available \href{https://cs.stanford.edu/\%7Ekeshik/halcyon}{here}.}
\endgroup

\newcommand{\methodname}[1]{\texttt{Halcyon}}

\newcommand{\keshik}[1]{%
  \begingroup
  \setlength{\fboxsep}{0.5pt}%
  \leavevmode
  \colorbox{orange}{\textcolor{white}{\strut Keshik}}%
  \textcolor{orange}{:\ #1}%
  \endgroup
}

\newcommand{\MP}[1]{%
  \begingroup
  \setlength{\fboxsep}{0.5pt}%
  \leavevmode
  \colorbox{blue}{\textcolor{white}{\strut MP}}%
  \textcolor{green}{:\ #1}%
  \endgroup
}

\newcommand{\yue}[1]{%
  \begingroup
  \setlength{\fboxsep}{0.5pt}%
  \leavevmode
  \colorbox{blue}{\textcolor{white}{\strut Yue}}%
  \textcolor{blue}{:\ #1}%
  \endgroup
}

\newcommand{\AT}[1]{%
  \begingroup
  \setlength{\fboxsep}{0.5pt}%
  \leavevmode
  \colorbox{purple}{\textcolor{white}{\strut AT}}%
  \textcolor{purple}{:\ #1}%
  \endgroup
}

\maketitle
\vspace{-1.2em}

\begin{abstract}
Modern text-to-image (T2I) systems use internet-scale language pretraining for 
text-conditioned image generation.
\textit{We ask whether language pretraining is necessary for strong T2I generation.}
We introduce \methodname{}, a family of early-fusion multimodal autoregressive models pretrained entirely from scratch on GPIC, a dataset of 100M image-text pairs.
\methodname{}-0.52B outperforms LlamaGen-XL/8 (\textbf{48.0} vs. {63.3} gFD) with \textbf{18.2$\times$} less pretraining compute.
Further, \methodname{}-1.6B outperforms JiT-T2I-XXL/8++ (\textbf{35.3} vs. {41.0} gFD) with \textbf{275$\times$} less pretraining compute.
Controlled experiments show that text-only pretraining does not improve T2I generation.
Halcyon also learns image captioning by replacing caption dropout with I2T supervision, maintaining competitive T2I generation quality under the same pretraining budget.
We show that strong T2I systems can be trained from scratch without internet-scale language pretraining.
\end{abstract}

\vspace{-0.5cm}
\begin{figure}[h]
\begin{adjustbox}{width=1\textwidth,center}
\begin{tabular}{c}
    \includegraphics[width=1\textwidth]
    {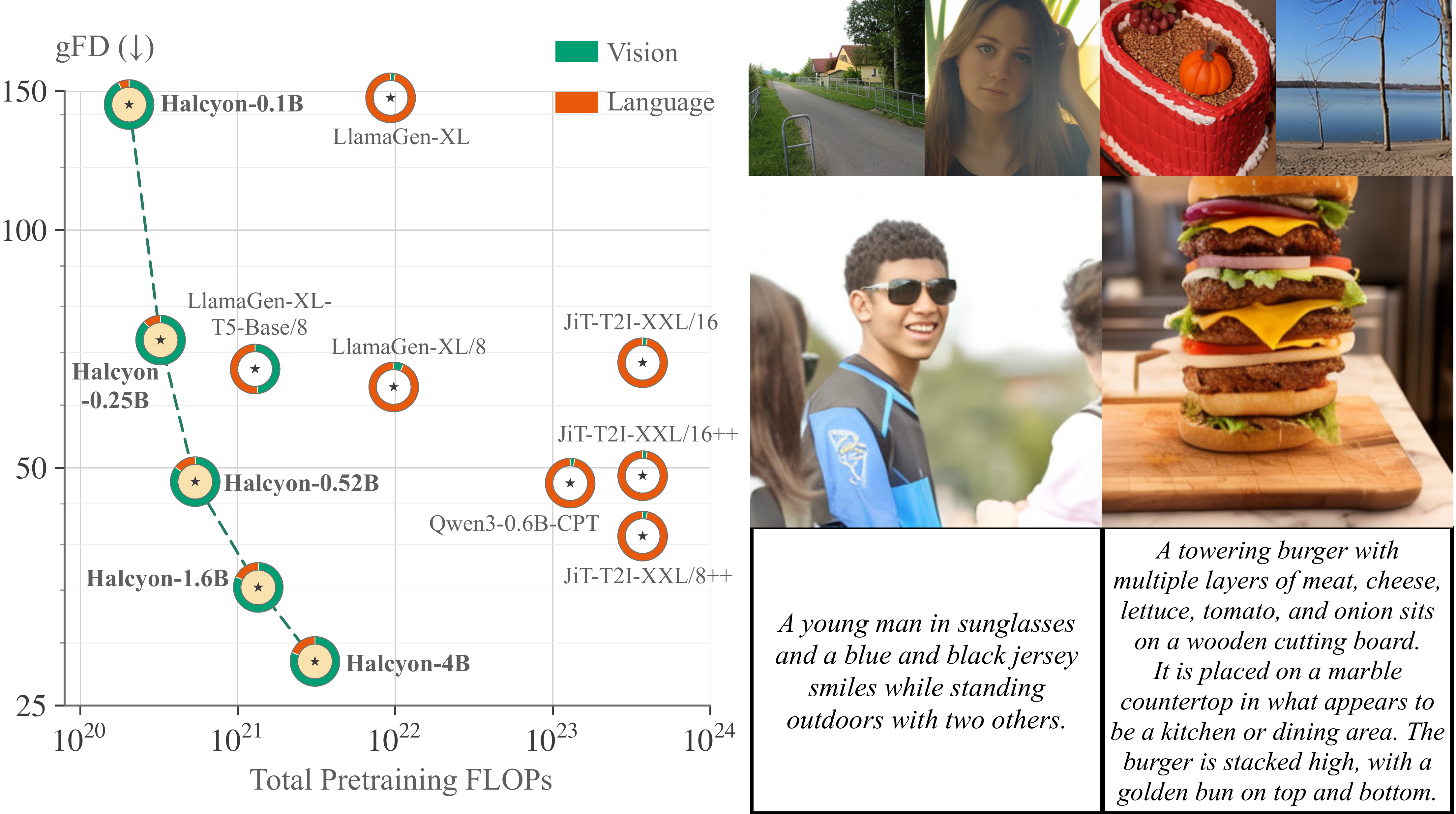} \\
\end{tabular}
\end{adjustbox}
\vspace{-0.4cm}
\caption{\textbf{Text-to-image generation without internet-scale language pretraining.}
\textbf{Left:} Generation quality on GPIC~\citep{chandrasegaran2026gpic} (gFD, log scale) vs.\ total pretraining FLOPs required to build each T2I system, including language model, image tokenizer, and image generator pretraining.
Each $\star$ marks total pretraining FLOPs and gFD, while the donut breaks down language and vision compute.
\textbf{\colorbox{mapit}{Halcyon (Ours)}, trained entirely from scratch, achieves lower gFD with substantially lower total pretraining compute} compared to:
\textit{(i) LlamaGen~\citep{sun2024autoregressive}}, late-fusion autoregressive models with 0.78--0.90B image generators and pretrained FLAN-T5 encoders (0.11--1.22B);
\textit{(ii) JiT-T2I~\citep{li2026back,ma2026pixelgen}}, late-fusion diffusion models with a 1.10B image generator and pretrained Qwen3-1.7B; and
\textit{(iii) Qwen3-0.6B-CPT}, an autoregressive model continually pretrained on image--text pairs, initialized from a pretrained Qwen3-0.6B language model~\citep{yang2025qwen3}.
\textbf{Right:} $256\times256$ samples generated by Halcyon-4B with prompts shown for two samples.
Best viewed in color.
}
\label{fig_main:compute_vs_gfd}
\vspace{-0.2cm}
\end{figure}

\input{sections/intro}
\input{sections/related_works}
\input{sections/method}
\input{sections/experiment_setup}
\input{sections/experiment_main_results}
\input{figures/samples_fig}
\input{sections/experiment_results_effect_of_language_pretraining}
\input{sections/experiment_results_effect_of_i2t_supervision}
\input{sections/conclusion}
\input{sections/acknowledgements}

\bibliography{iclr2027_conference}
\bibliographystyle{iclr2027_conference}

\clearpage
\newpage
\appendix
\startcontents[appendix]
\section*{Appendix}
\printcontents[appendix]{l}{1}[2]{}

\input{sections_supp/model_configs}
\input{tables/flops_vs_gfd}
\input{sections_supp/sampling_details}
\input{sections_supp/additional_results}
\input{sections_supp/training_seed_variance}
\input{sections_supp/sampling_seed_variance}
\input{sections_supp/inference_time}
\input{sections_supp/visualize}

\end{document}

%% file: math_commands.tex
\usepackage{amsmath,amsfonts,bm}

\def\Secref#1{Section~\ref{#1}}

\def\eqref#1{equation~\ref{#1}}

\def\1{\bm{1}}

\DeclareMathAlphabet{\mathsfit}{\encodingdefault}{\sfdefault}{m}{sl}
\SetMathAlphabet{\mathsfit}{bold}{\encodingdefault}{\sfdefault}{bx}{n}



%% file: sections/intro.tex
\section{Introduction}
\label{sec:intro}

Text-to-image (T2I) generation maps natural-language descriptions to images.
Modern T2I systems have an ingrained reliance on internet-scale language pretraining~\footnote{Throughout this paper, \emph{language pretraining} refers to pretraining on web-scale text-only corpora, excluding image-paired captions.} for semantic conditioning of image generators.
This dependence is realized via two designs, distinguished by how and when text tokens interact with image tokens:
(1)~\emph{Late fusion}, the predominant design, uses two separate models: a large  model is pretrained on text, and the resulting text embeddings are used to condition a separate image generator through cross- or joint attention~\citep{saharia2022photorealistic,chen2024pixart,flux2,zhao2026qwen};
(2)~\emph{Early fusion} removes the separate text encoder while retaining language pretraining within a single multimodal model trained on text-only and image-text data, with captions used as a prefix to condition image generation~\citep{team2024chameleon,wang2024emu3,zhou2025transfusion,liu2026tuna}.
Across both designs, language pretraining dominates the compute budget, accounting for over 80\% of total pretraining FLOPs in representative T2I systems~\citep{sun2024autoregressive,liu2026tuna}.
This raises two issues. First, its contribution to T2I generation has not been systematically quantified.
Second, language pretraining cost is often excluded, understating the total cost of building a T2I system.

In this work, we show that strong T2I generation can be achieved without language pretraining.
We introduce \methodname{}, a family of early-fusion autoregressive models that learn text conditioning directly from image--text pairs using next-token prediction.
We quantify the contribution of language pretraining to T2I generation using GPIC~\citep{chandrasegaran2026gpic}, a dataset of 100M image--text pairs, through three experimental axes: (1)~\textbf{From-scratch pretraining:} we train all components from scratch, including the text and image tokenizers and the generative model; (2)~\textbf{Data composition and scheduling:} under a fixed pretraining budget, we control the mixture and schedule of text-only, T2I, and I2T examples and measure their effect on T2I generation quality; and (3)~\textbf{Compute comparisons:} we pretrain representative T2I baselines on GPIC and compare generation quality as a function of total pretraining compute, decomposed into language and vision.
Through these experiments, we quantify the utility of language pretraining for T2I generation.

Our results show that 
\methodname{} achieves strong T2I generation without language pretraining, with generation Fréchet distance (gFD) consistently improving as model size scales (Figure~\ref{fig_main:compute_vs_gfd}).
Particularly, \methodname{}-0.52B achieves 48.0 gFD, outperforming LlamaGen-XL/8 (FLAN-T5-XL 1.22B + 0.90B image generator; 63.3 gFD) using \textbf{18.2$\times$ less} total pretraining compute.
\methodname{}-1.6B achieves 35.3 gFD, outperforming JiT-T2I-XXL/8++ (Qwen3-1.7B + 1.10B image generator; 41.0 gFD) using \textbf{275$\times$ less} total pretraining compute.
Our data composition and scheduling experiments further show that external text-only pretraining does not improve T2I generation.
Further, we\input{figures/related_work_scatter_plot}show that \methodname{} can learn image captioning by replacing caption-dropped T2I examples with I2T sequences, while maintaining competitive T2I generation quality under the same pretraining budget.
We provide a simple recipe for pretraining T2I models without language pretraining. Our work prompts rethinking the role of generic language pretraining for multimodal generation.

%% file: figures/related_work_scatter_plot.tex
\begin{wrapfigure}[16]{r}{0.5\linewidth}
\vspace{-15pt}
\centering
\begin{tikzpicture}[lbl/.style={font=\tiny, align=center, inner sep=1pt}]
\begin{loglogaxis}[
  width=\linewidth, height=0.9\linewidth,
  xmin=1e9, xmax=1e15, ymin=1e10, ymax=1e14,
  clip=false,
  xlabel={Language pretraining tokens},
  ylabel={Vision pretraining tokens},
  grid=major, grid style={gray!20},
  tick label style={font=\scriptsize},
  label style={font=\scriptsize},
  xlabel style={yshift=3pt}, ylabel style={yshift=-3pt},
  legend pos=north west,
  legend style={font=\tiny, draw=none, fill=none, row sep=-2pt},
  legend cell align=left,
]
\addplot[black, dashed, thin, forget plot]
  coordinates {(1e10,1e10) (1e14,1e14)};
\addplot[only marks, mark=*, mark size=1.8pt, color=red!85!black]
  coordinates {(1e12,6.9e12) (1e12,1.07e12) (3.6e13,1.21e11)};
\addplot[only marks, mark=square*, mark size=1.8pt, color=blue!70!black]
  coordinates {(2.9e12,1.9e12) (1e12,1e12) (1.71e12,4.59e12) (1.85e13,2.0e12)};
\addplot[only marks, mark=pentagon*, mark size=3.5pt, very thick, color=ourscoral]
  coordinates {(4.5e9,1.024e11)};
\legend{Late fusion, Early fusion, \methodname{} (ours)}
\node[lbl, anchor=south] at (axis cs:1.3e12,8.5e12)
  {LlamaGen-T2I\\\citep{sun2024autoregressive}};
\node[lbl, anchor=east]  at (axis cs:0.8e12,1.2e12)
  {PixArt-$\alpha$\\\citep{chen2024pixart}};
\node[lbl, anchor=north] at (axis cs:1.2e12,0.8e12)
  {Transfusion\\\citep{zhou2025transfusion}};
\node[lbl, anchor=north west] at (axis cs:1.8e12,1.75e12)
  {Chameleon\\\citep{team2024chameleon}};
\node[lbl, anchor=east] at (axis cs:1.45e12,4.4e12)
  {Emu3\\\citep{wang2024emu3}};
\node[lbl, anchor=west] at (axis cs:0.5e13,3.3e12)
  {Tuna-2\\\citep{liu2026tuna}};
\node[lbl, anchor=north] at (axis cs:3.6e13,0.95e11)
  {PixelGen\\\citep{ma2026pixelgen}};
\node[lbl, anchor=north, text=ourscoral] at (axis cs:4.5e9,0.85e11)
  {\methodname{}\\(Ours)};
\end{loglogaxis}
\end{tikzpicture}
\vspace{-25pt}
\caption{\textbf{Language vs.\ vision pretraining tokens.}
Language tokens count text-only pretraining. 
Emu3 and Tuna-2 values are estimated.
}
\label{fig:lang_vs_vision}
\vspace{-15pt}
\end{wrapfigure}
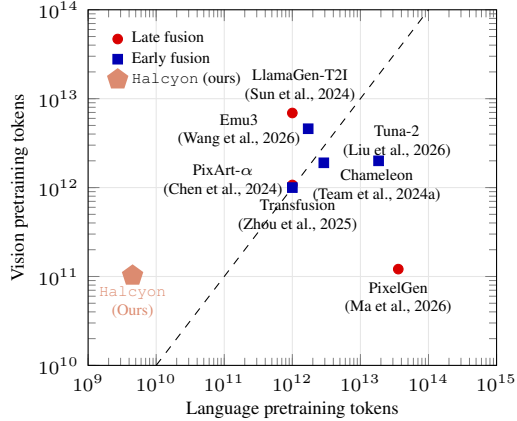

%% file: sections/related_works.tex
\section{Related Work}

Modern T2I systems use large-scale language pretraining through two designs. In \emph{late fusion}, the predominant paradigm, a language model is pretrained separately on text-only data, and its embeddings condition a visual generator. Imagen~\citep{saharia2022photorealistic}, Fluid~\citep{fan2025fluid} and PixArt-$\alpha$~\citep{chen2024pixart} use T5-XXL~\citep{chung2024scaling}, LlamaGen-T2I~\citep{sun2024autoregressive} uses T5-XL, SANA~\citep{xie2024sana} uses Gemma-2B~\citep{team2024gemma} and PixelGen~\citep{ma2026pixelgen} uses Qwen3-1.7B. 
In \emph{early fusion}, the separate text encoder is removed, but text-only pretraining is incorporated into multimodal pretraining. Chameleon~\citep{team2024chameleon}, Transfusion~\citep{zhou2025transfusion}, and Emu3~\citep{wang2024emu3} train jointly on text-only and image--text data, while Tuna-2~\citep{liu2026tuna} and SimpleAR~\citep{wang2025simplear} initialize from 
pretrained Qwen models.
Figure~\ref{fig:lang_vs_vision} compares the text and vision tokens used to pretrain these T2I systems. 
Systems with language pretraining consume $10^{12}\sim10^{14}$ text tokens, often exceeding the vision budget. In contrast, \methodname{} uses no text-only language pretraining; its only language supervision comes from caption tokens paired with images, more than two orders of magnitude lower than the language-pretraining budgets shown.
We note that early T2I systems such as DALL-E, CogView, and GLIDE learned text conditioning directly from paired image--text data, without language pretraining~\citep{ramesh2021zero,ding2021cogview,glide}.
However, the role of language pretraining has been debated: Imagen~\citep{saharia2022photorealistic} found benefits from scaling pretrained text encoders, whereas Parti~\citep{yu2022scaling} found that text-encoder pretraining only marginally reduced T2I generation loss.
Similarly, \cite{zhang2024pre} found limited benefits from initializing autoregressive T2I models with pretrained language models.
Nevertheless, modern T2I systems have largely shifted toward pretrained language models for text conditioning.
We quantify the role of language pretraining and show that T2I models trained entirely from scratch, without language pretraining, can outperform strong T2I baselines relying on language pretraining.

%% file: sections/method.tex
\section{Approach}
\label{sec:method}

\methodname{} is a decoder-only Transformer 
with a joint text--image vocabulary.
We model text and image tokens as a single sequence using next-token prediction, following prior multimodal early-fusion works
~\citep{team2024chameleon,wang2024emu3}.
All components, including the text tokenizer, image tokenizer, and
multimodal generative model are trained from scratch.
Below, we describe the tokenizers, sequence format, training objective,
and pretraining data composition and scheduling.

\vspace{-0.2cm}
\subsection{Tokenization}
\label{sec:tokenization}

\textbf{Images.}
We train a discrete image tokenizer on GPIC images, following the encoder--decoder design of Cosmos Tokenizer~\citep{agarwal2025cosmos} with finite scalar quantization (FSQ)~\citep{mentzer2024fsq}.
The tokenizer downsamples images by a factor of 8 along each spatial dimension, with a codebook size of 64{,}000.
Each $256\times256$ image is represented as
$N=1{,}024$ \imgname{image tokens}
$\mathbf{x}=(x_1,\ldots,x_N)$ obtained by flattening
the $32\times32$ grid in raster order.

\textbf{Text.}
We train a byte-level BPE tokenizer~\citep{sennrich2016bpe} from scratch on GPIC captions and OpenWebText~\cite{Gokaslan2019OpenWeb}, with a vocabulary size of $16{,}384$.
Each caption is represented as
\txtname{$\mathbf{c}=(c_1,\ldots,c_L)$}.

\textbf{Joint vocabulary.}
Text tokens, image tokens, and special tokens form a single vocabulary
of size $80{,}640$, comprising $16{,}384$ text tokens, $64{,}000$
image tokens, and $256$ special tokens.
The model uses a shared embedding table and output head.
We use \spctok{SOV} and \spctok{EOV} to mark the beginning and end
of an image, \spctok{EOS} to mark the end of a sequence, and
\spctok{PAD} for sequence padding.

\subsection{Sequence Format and Training Objective}
\label{sec:objective}

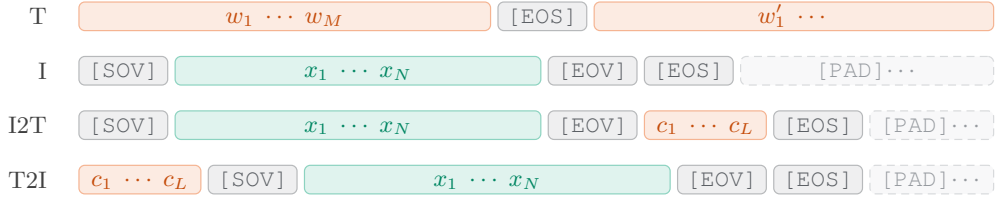
\begin{figure}[!t]
\centering
\begin{tikzpicture}[
  node distance=\chipgap,
  rowlabel/.style={font=\footnotesize, anchor=east, text=black!75},
]
\node[captionchip, anchor=west, minimum width=\txtw, font=\footnotesize] (w1) at (0,0) {$w_1\;\cdots\;w_M$};
\node[specialchip, right=of w1, font=\footnotesize] (q1) {\texttt{[EOS]}};
\bandfill{captionchip}{q1}{w1b}{$w'_1\;\cdots$}
\node[rowlabel] at ($(w1.west)+(-0.30,0)$) {$\mathrm{T}$};
 
\node[specialchip, anchor=west, font=\footnotesize] (s2) at (0,-0.72) {\texttt{[SOV]}};
\node[imagechip, right=of s2, minimum width=\imgw, font=\footnotesize] (x2) {$x_1\;\cdots\;x_N$};
\node[specialchip, right=of x2, font=\footnotesize] (e2) {\texttt{[EOV]}};
\node[specialchip, right=of e2, font=\footnotesize] (o2) {\texttt{[EOS]}};
\padfill{o2}{p2}
\node[rowlabel] at ($(s2.west)+(-0.30,0)$) {$\mathrm{I}$};
 
\node[specialchip, anchor=west, font=\footnotesize] (s3) at (0,-1.44) {\texttt{[SOV]}};
\node[imagechip, right=of s3, minimum width=\imgw, font=\footnotesize] (x3) {$x_1\;\cdots\;x_N$};
\node[specialchip, right=of x3, font=\footnotesize] (e3) {\texttt{[EOV]}};
\node[captionchip, right=of e3, minimum width=\capw, font=\footnotesize] (c3) {$c_1\;\cdots\;c_L$};
\node[specialchip, right=of c3, font=\footnotesize] (o3) {\texttt{[EOS]}};
\padfill{o3}{p3}
\node[rowlabel] at ($(s3.west)+(-0.30,0)$) {$\mathrm{I2T}$};
 
\node[captionchip, anchor=west, minimum width=\capw, font=\footnotesize] (c4) at (0,-2.16) {$c_1\;\cdots\;c_L$};
\node[specialchip, right=of c4, font=\footnotesize] (s4) {\texttt{[SOV]}};
\node[imagechip, right=of s4, minimum width=\imgw, font=\footnotesize] (x4) {$x_1\;\cdots\;x_N$};
\node[specialchip, right=of x4, font=\footnotesize] (e4) {\texttt{[EOV]}};
\node[specialchip, right=of e4, font=\footnotesize] (o4) {\texttt{[EOS]}};
\padfill{o4}{p4}
\node[rowlabel] at ($(c4.west)+(-0.30,0)$) {$\mathrm{T2I}$};
\end{tikzpicture}
\vspace{-0.3cm}
\caption{\textbf{Sequence formats.}
\txtname{Text-only pretraining tokens}, \txtname{caption tokens}, and \imgname{image tokens} occupy one joint vocabulary alongside special tokens.
The top two formats are unimodal, while the bottom two represent paired image-caption examples.
Rows $\mathrm{I}$ and $\mathrm{I2T}$ share the same \texttt{[SOV]}\,$\mathbf{x}$\,\texttt{[EOV]} prefix, which is also obtained when dropping the caption from a T2I sequence used for classifier-free guidance.
Image-caption sequences are padded to a fixed length, with \texttt{[PAD]} positions excluded from the training loss.
Box widths are schematic and not proportional to token counts.
}
\label{fig:sequences}
\end{figure}

We use four sequence formats for pretraining shown in Figure~\ref{fig:sequences}.
Text-only sequences are packed, while image-containing sequences
are padded to a fixed length with \padtok{} tokens. We do not pack multiple image--caption pairs
into a sequence.
For image-only (I) sequences, image tokens are predicted from
\spctok{SOV}.
For image-to-text (I2T) sequences, image tokens are
predicted first, followed by caption tokens conditioned on the image.
For text-to-image (T2I) sequences, image tokens are predicted
conditioned on the caption.
For classifier-free guidance (CFG)~\citep{ho2022classifier},
we adopt the standard practice of dropping the caption from a fraction
$p_{\mathrm{drop}}$ of T2I examples to train the unconditional branch.
At inference, the conditional and unconditional image-token logits are
combined as
$\boldsymbol{\ell}_{\mathrm{cfg}} =
\boldsymbol{\ell}_{\mathrm{u}} +
\gamma(\boldsymbol{\ell}_{\mathrm{c}} -
\boldsymbol{\ell}_{\mathrm{u}})$,
where $\gamma$ is the guidance scale.
We note that I2T sequences share the same image prefix
as unconditional T2I sequences, and therefore also train the
unconditional image branch used for CFG.
We use weighted next-token cross-entropy,
$\mathcal{L} =
(\mathcal{L}_{\mathrm{text}} +
\lambda_{\mathrm{img}}\mathcal{L}_{\mathrm{img}})/M$,
where
$\mathcal{L}_{\mathrm{text}} =
-\sum_{i\in\mathcal{T}}\log p_{\theta}(s_i\mid s_{<i})$,
$\mathcal{L}_{\mathrm{img}} =
-\sum_{i\in\mathcal{I}}\log p_{\theta}(s_i\mid s_{<i})$,
and $M=|\mathcal{T}|+|\mathcal{I}|$.
Here, $\mathcal{T}$ and $\mathcal{I}$ denote the non-padding text
and image tokens respectively, with special tokens
included in $\mathcal{T}$.
We set
$\lambda_{\mathrm{img}}=0.5$ to prevent image tokens dominating the training loss 
similar to Emu3~\citep{wang2024emu3}.

\subsection{Pretraining Data Composition and Scheduling}
\label{sec:mixtures}

To measure the utility of language pretraining for T2I generation, we  control the composition and scheduling of text-only, T2I, and I2T examples under a \textbf{fixed} pretraining budget. 
Figure~\ref{fig:mixture} illustrates the five configurations.
Unless otherwise specified, caption dropout is applied to T2I examples during training at rate $p_{\mathrm{drop}}$.
\begin{itemize}[noitemsep]
\item\textbf{Baseline T2I.}
We train exclusively on T2I examples.

\item\textbf{Staged: Language $\rightarrow$ T2I.}
We allocate the first $\alpha$ fraction of training to text-only pretraining, followed by T2I for the remaining $1-\alpha$ fraction.
We choose $\alpha$ such that the text-only pretraining tokens exceed the Chinchilla-optimal token budget for the model size~\citep{hoffmann2022chinchilla}.
This setup studies language-first pretraining schedules used in early-fusion systems such as Tuna-2~\citep{liu2026tuna}.
Separately, we evaluate continual pretraining recipe from an internet-scale pretrained LLM by expanding its vocabulary with image tokens and continuing training on paired image--text data.

\item\textbf{Staged: I2T $\rightarrow$ T2I.}
We allocate the first $\beta$ fraction of training to I2T, followed by T2I for the remaining $1-\beta$ fraction.

\item\textbf{Mixing: T2I + I2T.}
We interleave T2I and I2T examples throughout training, allocating a fraction $\eta$ to I2T and $1-\eta$ to T2I.

\item\textbf{Mixing: T2I without caption dropout + I2T.}
We set $\eta=p_{\mathrm{drop}}$ and disable caption dropout. This preserves the baseline's conditional and unconditional image-training tokens while additionally providing I2T supervision.
\end{itemize}

\input{figures/mixtures}

%% file: figures/mixtures.tex
%
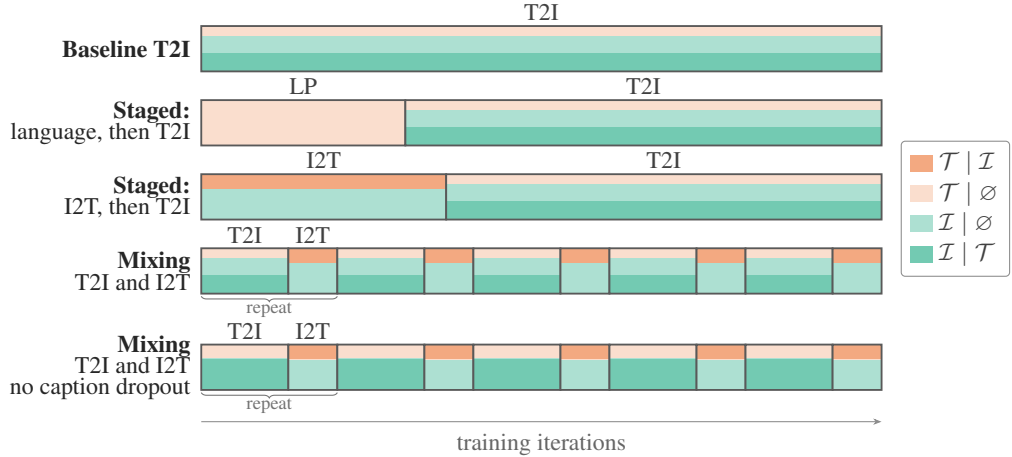
\begin{figure}[!b]
\vspace{-10pt}
\centering
\begin{tikzpicture}

\def\W{9.0}       
\def\RH{0.60}      
\def\RS{0.98}      
\def\RV{5.30}      
\def\GT{0}         

\newcommand{\phase}[8]{%
  \pgfmathsetmacro{\yb}{-#1*\RS}
  \pgfmathsetmacro{\xa}{(#2)*\W}\pgfmathsetmacro{\xb}{(#3)*\W}
  \pgfmathsetmacro{\ya}{\yb}
  \pgfmathsetmacro{\yA}{\yb+(#4)*\RH}
  \pgfmathsetmacro{\yB}{\yA+(#5)*\RH}
  \pgfmathsetmacro{\yC}{\yB+(#6)*\RH}
  \pgfmathsetmacro{\yD}{\yC+(#7)*\RH}
  \draw[tkcimg] (\xa,\ya) rectangle (\xb,\yA);
  \draw[tkuimg] (\xa,\yA) rectangle (\xb,\yB);
  \draw[tkutxt] (\xa,\yB) rectangle (\xb,\yC);
  \draw[tkctxt] (\xa,\yC) rectangle (\xb,\yD);
  \draw[black!65, line width=0.7pt, fill=none] (\xa,\yb) rectangle (\xb,{\yb+\RH});
  \node[phlab, anchor=south] at ({(\xa+\xb)/2},{\yb+\RH+0.04}) {#8};}

\newcommand{\rowname}[2]{%
  \pgfmathsetmacro{\yb}{-#1*\RS}
  \node[cfgname, anchor=east, inner sep=1pt] at (-0.1,{\yb+\RH/2}) {#2};}

\newcommand{\tilemarks}[1]{%
  \pgfmathsetmacro{\yb}{-#1*\RS}
  \node[phlab, anchor=south, inner sep=1pt] at ({0.064*\W},{\yb+\RH+0.03}) {T2I};
  \node[phlab, anchor=south, inner sep=1pt] at ({0.164*\W},{\yb+\RH+0.03}) {I2T};
  \draw[decorate, decoration={brace, mirror, amplitude=2pt, raise=1pt},
        black!55, line width=0.4pt] (0,\yb) -- ({0.20*\W},\yb)
        node[midway, below=2.5pt, inner sep=0.5pt, font=\scriptsize, text=black!55] {repeat};}

\rowname{1}{\textbf{Baseline T2I}}
\phase{1}{0}{1}{0.42}{0.36}{0.22}{0}{T2I}

\rowname{2}{\textbf{Staged:}\\[-2pt]language, then T2I}
\phase{2}{0}{0.30}{0}{0}{1}{0}{LP}
\phase{2}{0.30+\GT}{1}{0.42}{0.36}{0.22}{0}{T2I}

\rowname{3}{\textbf{Staged:}\\[-2pt]I2T, then T2I}
\phase{3}{0}{0.36}{0}{0.68}{0}{0.32}{I2T}
\phase{3}{0.36+\GT}{1}{0.42}{0.36}{0.22}{0}{T2I}

\rowname{4}{\textbf{Mixing}\\[-2pt]T2I and I2T}
\foreach \k in {0,...,4}{
  \pgfmathsetmacro{\s}{\k*0.20}
  \phase{4}{\s}{\s+0.128}{0.42}{0.36}{0.22}{0}{}
  \phase{4}{\s+0.128+\GT}{\s+0.20-\GT}{0}{0.68}{0}{0.32}{}}
\tilemarks{4}

\rowname{\RV}{\textbf{Mixing}\\[-2pt]T2I and I2T\\[-2pt]no caption dropout}
\foreach \k in {0,...,4}{
  \pgfmathsetmacro{\s}{\k*0.20}
  \phase{\RV}{\s}{\s+0.128}{0.70}{0}{0.30}{0}{}
  \phase{\RV}{\s+0.128+\GT}{\s+0.20-\GT}{0}{0.68}{0}{0.32}{}}

\tilemarks{\RV}

\pgfmathsetmacro{\yl}{-\RV*\RS-0.42}
\draw[-{Stealth[length=3.2pt]}, black!45, line width=0.4pt]
  (0,\yl) -- ({\W},\yl);
\node[axlab, anchor=north] at ({\W/2},{\yl-0.04}) {training iterations};

\pgfmathsetmacro{\kx}{\W+0.38}                                
\pgfmathsetmacro{\kytop}{-(1+\RV)/2*\RS + \RH/2 + 0.48}       
\newcommand{\keyitem}[3]{
  \pgfmathsetmacro{\yk}{\kytop-#1*0.40}
  \draw[#2] (\kx,\yk) rectangle ({\kx+0.30},{\yk+0.22});
  \node[font=\footnotesize, text=black!75, anchor=west, inner sep=1pt] at ({\kx+0.34},{\yk+0.11}) {#3};}

\begin{scope}[local bounding box=key]
  \keyitem{0}{tkctxt}{$\mathcal{T}\mid\mathcal{I}$}
  \keyitem{1}{tkutxt}{$\mathcal{T}\mid\varnothing$}
  \keyitem{2}{tkuimg}{$\mathcal{I}\mid\varnothing$}
  \keyitem{3}{tkcimg}{$\mathcal{I}\mid\mathcal{T}$}
\end{scope}
\draw[black!45, line width=0.4pt, rounded corners=1.5pt]
  ([shift={(-0.12,-0.10)}]key.south west) rectangle ([shift={(0.12,0.10)}]key.north east);

\end{tikzpicture}
\vspace{-10pt}
\caption{\textbf{Pretraining data composition and scheduling described in \Secref{sec:mixtures}.}
The four token types are:
\legsw{tkcimg}~image tokens conditioned on text ($\mathcal{I}\mid\mathcal{T}$);
\legsw{tkuimg}~image tokens without text conditioning ($\mathcal{I}\mid\varnothing$);
\legsw{tkctxt}~text tokens conditioned on images ($\mathcal{T}\mid\mathcal{I}$);
and \legsw{tkutxt}~text-only tokens ($\mathcal{T}\mid\varnothing$).
Vertical boundaries denote training phases, while horizontal bands
show the token composition within each phase.
Bands represent interleaved examples.
Band height and width are schematic. 
}
\label{fig:mixture}
\end{figure}

%% file: sections/experiment_setup.tex
\section{Experiments}
\label{main:exp}

\subsection{Experimental Setup}
\vspace{-0.07cm}
\textbf{Dataset and pretraining protocol.}
We pretrain on GPIC~\citep{chandrasegaran2026gpic}, a dataset of approximately 100M image--text pairs, at $256 \times 256$ resolution.
All configurations are trained for one epoch on GPIC with a global batch size of 1,024 and an identical warmup-stable-decay (WSD) learning-rate schedule~\citep{hu2024minicpm}.
For language pretraining, we use ClimbMix~\citep{climbmix} packed without padding.
Model details and training details are provided in Appendix~\ref{sec:supp:model}.

\textbf{Tokenizer training and evaluation.}
$\bullet$ \textit{Image tokenizer.}
We train our image tokenizer for one epoch using reconstruction and perceptual losses, followed by an additional half epoch with adversarial loss\input{tables/image_tokenizer_results}using a PatchGAN discriminator.
Compared to Cosmos-DI-8x8~\citep{agarwal2025cosmos}, we increase the decoder width from 128 to 192.
To assess reconstruction quality, we compare our tokenizer with pretrained LlamaGen-VQ-16x16 and Cosmos-DI-8x8 tokenizers. 
We include Real-vs-Real scores as a reference.
Our tokenizer outperforms Cosmos-DI-8x8 across all metrics (Table~\ref{tab:tokenizer}).
As with prior discrete tokenizers, reconstruction fidelity is lower for small text and fine glyphs~\citep{team2024chameleon,sun2024autoregressive,agarwal2025cosmos}.

$\bullet$ \textit{Text tokenizer.}
Despite having a $9.3\times$ smaller vocabulary than Qwen3, our tokenizer achieves comparable compression on GPIC test captions (4.774 vs.\ 4.763 bytes/token), outperforming the FLAN-T5 tokenizer (4.152 bytes/token; Table~\ref{tab:text_tokenizer}).
On held-out OpenWebText, our tokenizer achieves 4.086\input{tables/bpe_tokenizer_results}bytes/token.
OpenWebText is used only to train the BPE tokenizer, not the multimodal generative models.
Full tokenizer training details are provided in Table~\ref{tab:tokenizer_configs}.

\textbf{Model configurations.}
We train \methodname{} at five sizes: 0.1B, 0.25B, 0.52B, 1.6B, and 4B parameters.
Model architecture details are provided in Table~\ref{tab:configs}. 
We use a sequence length of 1280 for all experiments.
We use $p_{drop}=0.1$ unless otherwise specified.

\textbf{Late-fusion methods.}
Late fusion is the predominant design choice for modern T2I systems.
Selecting baselines for this study is non-trivial: existing systems differ in pretraining objective, language conditioning, image representation, and training setup, while our study requires public pretraining code and models that enable controlled interventions to study the role of language pretraining.
We select two representative families, LlamaGen~\citep{sun2024autoregressive} and JiT-T2I~\citep{li2026back,ma2026pixelgen}.
They cover (i) autoregressive and diffusion-based generation,
(ii) encoder-decoder and decoder-only language models of different sizes,
and (iii) image sequence lengths of 256 and 1{,}024.
For both families, we use the authors' official codebases and follow their training configurations unless stated otherwise.
Below, we describe our setup and summarize them in Table~\ref{tab:baselines}.

$\bullet$ \textbf{LlamaGen~\citep{sun2024autoregressive}.}
This method uses a FLAN-T5-XL text encoder for language modeling, a discrete image tokenizer, and an autoregressive image generator.
\begin{itemize}[noitemsep]
    \item \textbf{Default (LlamaGen-XL).}
    We use the original LlamaGen configuration: FLAN-T5-XL, the pretrained image tokenizer producing 256 image tokens from a codebook of 16{,}384 entries, and the 0.78B image generator.

    \item \textbf{Use \methodname{} image tokenizer (LlamaGen-XL/8).}
    Starting with the default configuration, we replace the image tokenizer with our GPIC-trained tokenizer.
    This increases the visual vocabulary size by $4\times$, the image sequence length from 256 to 1{,}024, and the image generator size from 0.78B to 0.90B.
    
    \item \textbf{Use smaller text encoder (LlamaGen-XL-T5-Base/8).}
    Starting with LlamaGen-XL/8, we replace FLAN-T5-XL~(1.22B) with FLAN-T5-base~(0.11B) to study the effect of language model size.
\end{itemize}

$\bullet$ \textbf{JiT-T2I~\citep{li2026back,ma2026pixelgen}.}
This method uses pretrained Qwen3-1.7B~\citep{yang2025qwen3} as the language model and a pixel-space flow-matching image generator.
\begin{itemize}[noitemsep]
    \item \textbf{Default (JiT-T2I-XXL/16).}
    We use the original JiT-T2I-XXL/16 configuration: Qwen3-1.7B and no image tokenizer, with $16 \times 16$ image patches producing 256 image tokens.

    \item \textbf{Add perceptual loss (JiT-T2I-XXL/16 ++).}
    Starting with the default configuration, we add LPIPS loss~\citep{zhang2018unreasonable} to train the image generator.
    This tests the effect of perceptual supervision used to also train the image tokenizers in \methodname{} and LlamaGen.

    \item \textbf{Use $8 \times 8$ patches (JiT-T2I-XXL/8 ++).}
    Starting with JiT-T2I-XXL/16 ++, we reduce the image patch size from $16 \times 16$ to $8 \times 8$, increasing the image sequence length from 256 to 1{,}024, matching \methodname{} and the modified LlamaGen configurations.
\end{itemize}

\textbf{Early-fusion methods.}
Existing early-fusion models, including Chameleon~\citep{team2024chameleon} and Emu3~\citep{wang2024emu3}, incorporate language pretraining into multimodal generative modeling.
We study the role of language pretraining in two controlled settings.
First, we vary the composition and scheduling of text-only, T2I, and I2T sequences within \methodname{} under a fixed pretraining budget (Section~\ref{sec:mixtures}).
Second, we study continual pretraining by initializing from Qwen3-0.6B-Base, pretrained on 36T text tokens, expanding its vocabulary with image tokens, and continuing pretraining on GPIC using the baseline T2I recipe.
For comparison, we train the same architecture and vocabulary from scratch under an identical T2I pretraining budget.
We disable weight decay for embedding parameters when vocabulary size is large (216K).

\begin{table}[t]
\centering
\caption{\textbf{Experiment configurations for late-fusion methods.}
LlamaGen and JiT-T2I configurations, varying the language model,
image sequence length, and perceptual supervision.
LlamaGen and \methodname{} are autoregressive; JiT is diffusion-based.
Pretraining tokens report the total language and vision tokens used during training; \methodname{} uses no pretrained language model, and all its language tokens come from GPIC captions.
We report encoder parameters for FLAN-T5.
}
\label{tab:baselines}
\small
\setlength{\tabcolsep}{8pt}
\adjustbox{max width=\textwidth}{
\begin{tabular}{@{}l c cc rr c@{}}
\toprule
& Language
& \multicolumn{2}{c}{Params}
& \multicolumn{2}{c}{Pretraining tokens}
& Perceptual \\
\cmidrule(lr){3-4}
\cmidrule(lr){5-6}
Configuration
& model
& Language & Vision
& Language & Vision
& loss \\
\midrule
LlamaGen-XL                  & FLAN-T5-XL   & 1.22B & 0.78B & $\sim$1T & 25.6B  & \checkmark \\  
LlamaGen-XL/8
& FLAN-T5-XL   & 1.22B & 0.90B & $\sim$1T & 102.4B & \checkmark \\
LlamaGen-XL-T5-Base/8
& FLAN-T5-Base & 0.11B & 0.90B & $\sim$1T & 102.4B & \checkmark \\
\midrule
JiT-T2I-XXL/16               & Qwen3        & 1.70B & 1.10B & 36T      & 25.6B  &            \\
JiT-T2I-XXL/16 ++
& Qwen3        & 1.70B & 1.10B & 36T      & 25.6B  & \checkmark \\
JiT-T2I-XXL/8 ++
& Qwen3        & 1.70B & 1.10B & 36T      & 102.4B & \checkmark \\
\midrule
\multicolumn{1}{@{}>{\columncolor{mapit}[0pt][\tabcolsep]}l}{\textbf{\methodname{}~(Ours)}}
  & \cellcolor{mapit}\textemdash
  & \multicolumn{2}{>{\columncolor{mapit}}c}{0.1--4B}
  & \cellcolor{mapit}4.5B
  & \cellcolor{mapit}102.4B
  & \multicolumn{1}{>{\columncolor{mapit}[\tabcolsep][0pt]}c@{}}{\checkmark} \\
\bottomrule
\end{tabular}
}
\vspace{-10pt}
\end{table}

\textbf{Evaluation protocol.}
We evaluate all models at $256 \times 256$ resolution on three benchmarks:
$\bullet$~\textbf{GPIC (primary metrics).}
We follow the GPIC evaluation protocol using the GPIC Test-50K captions\footnote{\href{https://hf.co/datasets/stanford-vision-lab/gpic/blob/main/gpic_test_50k.jsonl}{\nolinkurl{hf.co/datasets/stanford-vision-lab/gpic/blob/main/gpic_test_50k.jsonl}}}, reporting generation Fréchet distance (gFD), Precision, Recall, Density, Coverage, and MMD.
$\bullet$~\textbf{GenEval~\citep{ghosh2023geneval}.}
We evaluate compositional generation using the rephrased prompts from PixelGen~\citep{ma2026pixelgen}\footnote{\href{https://github.com/Zehong-Ma/PixelGen/blob/main/evaluations/geneval/}{\nolinkurl{github.com/Zehong-Ma/PixelGen/blob/main/evaluations/geneval/}}}, removing generic phrases such as ``an image of'' and ``the image depicts''.
$\bullet$~\textbf{DPG-Bench~\citep{hu2024ella}.}
We evaluate complex prompt adherence on DPG-Bench using the original prompts.
$\bullet$ \textbf{Token-level NLL.}
For \methodname{}, we report negative log-likelihood (NLL) on GPIC Test-50K for four settings: text-conditioned images ($\mathcal{I}\mid\mathcal{T}$), unconditional images ($\mathcal{I}\mid\varnothing$), image-conditioned text ($\mathcal{T}\mid\mathcal{I}$), and unconditional text ($\mathcal{T}\mid\varnothing$).
$\bullet$ \textbf{Image captioning.}
We evaluate image--text alignment using Qwen3-VL-Reranker-8B scores over GPIC Test-50K images and generated captions, treating the image as the query and the caption as the document.

For each model, we sweep guidance scales on GPIC Validation-50K\footnote{\href{https://hf.co/datasets/stanford-vision-lab/gpic/blob/main/gpic_val_50k.jsonl}{\nolinkurl{hf.co/datasets/stanford-vision-lab/gpic/blob/main/gpic_val_50k.jsonl}}} and use the scale achieving the lowest gFD for all three benchmarks.
We report the sampling and guidance settings for each model in Appendix~\ref{sec:supp:model}.

%% file: tables/image_tokenizer_results.tex
\begin{wraptable}{r}{0.55\textwidth}
    \centering
    \vspace{-1.5em}
    \caption{
    \textbf{Reconstruction results on GPIC Test-50K ($256 \times 256$).} Our tokenizer, trained from scratch on GPIC, outperforms Cosmos-DI-8x8 on all metrics.
    }
    \label{tab:tokenizer}
    \resizebox{\linewidth}{!}{
    \begin{tabular}{lrccc}
        \toprule
        Tokenizer
        & rFD $\downarrow$
        & PSNR $\uparrow$
        & SSIM $\uparrow$
        & LPIPS $\downarrow$ \\
        \midrule
        LlamaGen-VQ-16x16
        &129.87  & 20.18 & 0.560  & 0.146  \\
        Cosmos-DI-8x8
        & 37.93 & 25.38 & 0.776 & 0.153 \\
        \rowcolor{mapit}
        \methodname~
        & \textbf{18.86}
        & \textbf{25.61}
        & \textbf{0.792}
        & \textbf{0.062} \\
        \midrule
        \rowcolor{gray!15}
        Real-vs-Real
        & 7.59 & -- & -- & -- \\
        \bottomrule
    \end{tabular}
    }
    \vspace{-1.1em}
\end{wraptable}

%% file: tables/bpe_tokenizer_results.tex
\begin{wraptable}[8]{r}{0.52\textwidth}
\vspace{-18pt}
\centering
\caption{\textbf{Text tokenizer comparison on GPIC test captions.}
Higher bytes/token and lower tokens/caption indicate better compression.}
\label{tab:text_tokenizer}
\small
\setlength{\tabcolsep}{4pt}
\begin{tabular}{lrcc}
\toprule
Tokenizer & Vocab. & Bytes/token $\uparrow$ & Tok./caption $\downarrow$ \\
\midrule
FLAN-T5    & 32,100  & 4.152 & 51.1 \\
Qwen3      & 151,669 & 4.763 & 44.6 \\
\rowcolor{mapit}
\methodname~       & \textbf{16,384}  & \textbf{4.774} & \textbf{44.5} \\
\bottomrule
\end{tabular}
\vspace{-8pt}
\end{wraptable}

%% file: sections/experiment_main_results.tex
\newpage
\subsection{Results: Comparison with Late-fusion T2I Methods}
\label{sec:baseline_comparison}

We report quantitative results in Table~\ref{tab:t2i_baselines}, including compositional generation benchmarks (GenEval and DPG-Bench) in Table~\ref{tab:t2i_geneval_dpg}.
We include generated samples in Figure~\ref{fig:generated_examples} and Appendix~\ref{sec:supp:visualize}.
Figure~\ref{fig_main:compute_vs_gfd} compares generation quality as a function of total pretraining FLOPs, with the exact compute breakdown reported in Table~\ref{tab:pretraining_compute}.
Below, we discuss our key findings:

\input{tables/results_with_late_fusion_baselines}

\noindent\textbf{Effect of image representation and perceptual supervision.}
For LlamaGen-XL, replacing its original image tokenizer with our GPIC-trained tokenizer increases the image sequence length from 256 to 1{,}024 and reduces gFD (147.0 $\rightarrow$ 63.3).
For JiT-T2I-XXL/16, adding LPIPS perceptual supervision reduces gFD (67.9 $\rightarrow$ 48.9).
Reducing the image patch size from $16\times16$ to $8\times8$ increases the image sequence length from 256 to 1{,}024 and further reduces gFD (48.9 $\rightarrow$ 41.0).
We use LlamaGen-XL/8 and JiT-T2I-XXL/8++ for subsequent comparisons with \methodname{}.

\noindent\textbf{Effect of language encoder size.}
We study whether a larger pretrained language encoder improves\input{tables/results_with_late_fusion_baselines_geneval_dpg}T2I generation.
For LlamaGen, replacing FLAN-T5-XL with FLAN-T5-base reduces the language encoder size from 1.22B to 0.11B while retaining the same image tokenizer and 0.90B image generator.
The resulting gFD changes only marginally (63.3 $\rightarrow$ 66.7), despite a 13.5$\times$ reduction in language-model training compute\footnote{Pretraining and finetuning the full FLAN-T5-XL encoder--decoder model requires $9.06\times10^{21}$ FLOPs vs.\ $6.69\times10^{20}$ FLOPs for FLAN-T5-base~\citep{chung2024scaling}}.
This provides initial evidence that an order-of-magnitude increase in language pretraining compute yields only marginal improvements in T2I generation.
We next investigate whether strong T2I generation can be achieved without language pretraining altogether.

\noindent\textbf{T2I generation without language pretraining.}
We now address the central question of this work: \textit{can a T2I model trained entirely from scratch match or surpass strong late-fusion systems that rely on pretrained language models?}
\methodname{}-0.52B, a single multimodal model trained entirely from scratch, achieves 48.0 gFD, outperforming LlamaGen-XL/8 with pretrained FLAN-T5-XL (63.3) and approaching JiT-T2I-XXL/8++ with pretrained Qwen3-1.7B (41.0).
Increasing \methodname{} to 1.6B parameters further reduces gFD to 35.3, surpassing both late-fusion methods.
At 4B parameters, gFD decreases to 28.4.
On compositional generation benchmarks, \methodname{}-0.52B  outperforms JiT-T2I-XXL/8++ on GenEval (44.9 vs.\ 43.2) and matches its DPG-Bench score (76.84 vs.\ 76.85).
At 1.6B, \methodname{} surpasses JiT-T2I-XXL/8++ on both GenEval (53.5 vs.\ 43.2) and DPG-Bench (77.9 vs.\ 76.9).
Moreover, \methodname{}-1.6B achieves these results with 275$\times$ less total pretraining compute than JiT-T2I-XXL/8++.
Together, these results demonstrate that strong T2I generation capabilities can be learned from paired image--text data alone, without language pretraining, while substantially reducing total pretraining compute.

\begin{takeaway}
\textbf{Takeaway.}
\methodname{} demonstrates that T2I models trained entirely from scratch, without internet-scale language pretraining, can outperform strong late-fusion methods while requiring substantially less total pretraining compute.
\end{takeaway}

%% file: tables/results_with_late_fusion_baselines.tex
\vspace{-0.5cm}
\begin{table*}[!h]
\centering
\caption{\textbf{Comparison of \methodname{} with late-fusion T2I methods on GPIC.}
Without language pretraining, \methodname{} achieves stronger T2I generation than the late-fusion baselines: \methodname{}-0.52B outperforms LlamaGen-XL/8 (48.0 vs.\ 63.3 gFD), while \methodname{}-1.60B outperforms JiT-T2I-XXL/8++ (35.3 vs.\ 41.0 gFD).
Vision parameter counts refer to the generative models, excluding image tokenizers (72M for LlamaGen and 136M for \methodname{}).
The CFG scale for each model is reported in Appendix Table~\ref{tab:t2i_cfg}.
}
\label{tab:t2i_baselines}
\small
\setlength{\tabcolsep}{3.2pt}

\adjustbox{max width=\textwidth}{
\begin{tabular}{l cc rccccc}
\toprule
Model
& Language
& Vision
& gFD $\downarrow$
& Prec. $\uparrow$
& Rec. $\uparrow$
& Dens. $\to$ 1
& Cov. $\uparrow$
& MMD $\downarrow$ \\
\midrule

LlamaGen-XL
& 1.22B & 0.78B
&147.01	&0.914	&0.715	&0.925	&0.822	&0.174\\

LlamaGen-XL/8
& 1.22B & 0.90B
&63.29	&0.948	&0.827	&1.058	&0.925	&0.041\\

LlamaGen-XL-T5-Base/8
& 0.11B & 0.90B
&66.69	&0.941	&0.822	&1.016	&0.924	&0.045\\

\midrule

JiT-T2I-XXL/16
& 1.70B & 1.10B
&67.92	&0.934	&0.822	&0.979	&0.923	&0.038\\

JiT-T2I-XXL/16 ++
& 1.70B & 1.10B
&48.86	&0.947	&0.861	&1.022	&0.939	&0.022\\

JiT-T2I-XXL/8 ++
& 1.70B & 1.10B
&40.97	&0.949	&0.880	&1.049	&0.952	&0.016\\

\midrule

\rowcolor{mapit!30}
& \multicolumn{2}{>{\columncolor{mapit!20}}c}{0.10B}
& 144.16 & 0.924 & 0.636 & 1.029 & 0.853 & 0.131 \\
\rowcolor{mapit!45}
& \multicolumn{2}{>{\columncolor{mapit!40}}c}{0.25B}
& 72.55 & 0.946 & 0.796 & 1.050 & 0.922 & 0.050 \\

\rowcolor{mapit!60}
& \multicolumn{2}{>{\columncolor{mapit!60}}c}{0.52B}
& 48.02 & 0.953 & 0.858 & 1.064 & 0.938 & 0.027 \\

\rowcolor{mapit!80}
& \multicolumn{2}{>{\columncolor{mapit!80}}c}{1.60B}
& 35.28 & 0.955 & 0.891 & 1.071 & 0.955 & 0.019 \\

\rowcolor{mapit}
\multirow{-5}{*}{\textbf{\methodname{}~(Ours)}}
& \multicolumn{2}{>{\columncolor{mapit}}c}{4.00B}
& {28.43} & 0.951 & 0.905 & 1.034 & 0.960 & {0.014} \\

\bottomrule
\end{tabular}
}
\end{table*}

%% file: tables/results_with_late_fusion_baselines_geneval_dpg.tex
\begin{wraptable}[17]{r}{0.55\columnwidth}
\centering
\vspace{-0.55cm}
\caption{\textbf{Comparison of \methodname{} with late-fusion T2I methods on GenEval and DPG-Bench.}
We also report GPIC gFD for comparison.
Per-category breakdowns of GenEval and DPG-Bench are provided in Sec.~\ref{sec:supp_geneval_breakdown}.}
\label{tab:t2i_geneval_dpg}
\small
\setlength{\tabcolsep}{3.2pt}
\adjustbox{max width=\linewidth}{%
\begin{tabular}{l rcc}
\toprule
Model
& GPIC gFD $\downarrow$
& GenEval $\uparrow$
& DPG-Bench $\uparrow$ \\
\midrule

LlamaGen-XL
& 147.01 & 30.03 & 65.40 \\

LlamaGen-XL/8
& 63.29 & 30.48 & 67.08 \\

LlamaGen-XL-T5-Base/8
& 66.69 & 28.48 & 64.67 \\

\midrule

JiT-T2I-XXL/16
& 67.92 & 36.27 & 72.69 \\

JiT-T2I-XXL/16 ++
& 48.86 & 41.79 & 76.58 \\

JiT-T2I-XXL/8 ++
& 40.97 & 43.17 & 76.85 \\

\midrule

\rowcolor{mapit!30}
\methodname{}-0.10B
& 144.16 & 27.31 & 66.36 \\

\rowcolor{mapit!45}
\methodname{}-0.25B
& 72.55 & 35.72 & 72.66 \\

\rowcolor{mapit!60}
\methodname{}-0.52B
& 48.02 & 44.91 & 76.84 \\

\rowcolor{mapit!80}
\methodname{}-1.60B
& 35.28 & 53.48 & 77.90 \\

\rowcolor{mapit}
\methodname{}-4.00B
& 28.43 & 56.39 & 77.90 \\

\bottomrule
\end{tabular}
}
\end{wraptable}

%% file: figures/samples_fig.tex
\vspace{0.5cm}
\begin{figure}[!h]
    \centering
    \includegraphics[width=\linewidth]{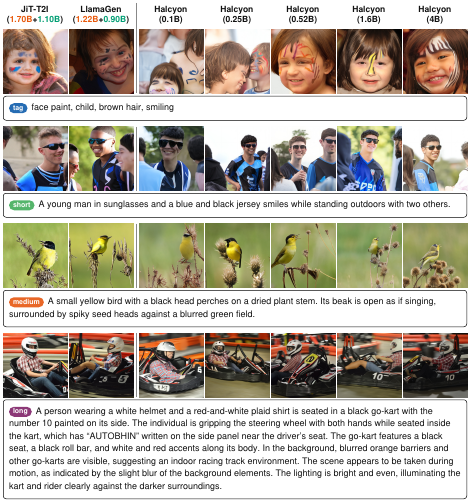}
    \vspace{-20pt}
    \caption{
    \textbf{Comparison of generated images.}
    From left to right: JiT-T2I-XXL/8++, LlamaGen-XL/8, and Halcyon models from 0.1B to 4B parameters.
    For the baselines, we report the sizes of the {\color{orange}\bf language model} and {\color{darkgreen}\bf image generator} separately.
    Image quality improves as \methodname{} scales up.
    All prompts are from GPIC Test-50K.
    Additional samples are provided in Appendix~\ref{sec:supp:visualize}.
    }
    \label{fig:generated_examples}
\end{figure}

%% file: sections/experiment_results_effect_of_language_pretraining.tex
\newpage
\subsection{Results: Effect of Language Pretraining on T2I generation}

\noindent\textbf{External language pretraining does not improve T2I generation.}
We investigate this through two controlled interventions.
First, under a fixed pretraining budget, allocating $\alpha{=}0.1, 0.2, 0.5$ to text-only pretraining reduces paired image--text training and worsens gFD from 48.0 to 51.6, 52.8, and 65.1, respectively; conditional image-token NLL also increases (Table~\ref{tab:language_pretraining}).
To isolate the effect of language pretraining from reduced T2I training, we additionally evaluate language pretraining followed by a full T2I epoch ($1.5\times$ total compute), but gFD does not improve (48.0 vs.\ 49.2).
Second, we investigate continual pretraining from Qwen3-0.6B-Base, pretrained on 36T text tokens.
Despite this extensive language pretraining, Qwen3-0.6B-CPT performs similar to the same architecture trained from random initialization (47.8 vs.\ 46.6 gFD).
Reducing the peak learning rate from $7.5\times10^{-4}$ to $1\times10^{-4}$ for continual pretraining also fails to improve T2I generation (Table~\ref{tab:qwen_init}).
Together, these experiments show that external text-only pretraining provides no benefit to T2I generation under the evaluated settings.

\input{tables/lp_effect}
\input{tables/continual_pretraining_results}

%% file: tables/lp_effect.tex
\begin{table*}[!h]
\centering
\caption{\textbf{Effect of language pretraining on \methodname{}-0.52B.}
Under a fixed pretraining budget, we measure the impact of allocating compute to external text-only pretraining on T2I generation.
Each language-pretraining stage exceeds the Chinchilla reference of 20 text tokens per parameter.
The $^\dagger$ configuration preserves a full T2I epoch after additional language pretraining ($1.5\times$ total pretraining compute), but does not improve gFD.
Token counts exclude padding.
}
\label{tab:language_pretraining}
\vspace{4pt}
\small
\setlength{\tabcolsep}{5pt}

\resizebox{\textwidth}{!}{%
\begin{tabular}{@{}l c rc rc cccc c@{}}
\toprule
& &
\multicolumn{2}{c}{Text tokens (B)}
& \multicolumn{2}{c}{Image tokens (B)}
& \multicolumn{4}{c}{NLL $\downarrow$}
& \\
\cmidrule(lr){3-4}
\cmidrule(lr){5-6}
\cmidrule(lr){7-10}
Configuration
& $\alpha$
& $\mathcal{T}\mid\varnothing$
& $\mathcal{T}\mid\mathcal{I}$
& $\mathcal{I}\mid\varnothing$
& $\mathcal{I}\mid\mathcal{T}$

& $\mathcal{T}\mid\varnothing$
& $\mathcal{T}\mid\mathcal{I}$
& $\mathcal{I}\mid\varnothing$
& $\mathcal{I}\mid\mathcal{T}$
& gFD $\downarrow$ \\
\midrule

\methodname{}-0.52B
& --
& 4.05 & 0
& 10.24 & 92.16
& \textbf{1.433} & \textbf{18.414} & \textbf{8.428} & \textbf{8.366}
& \textbf{48.02} \\

\midrule

Staged: Language $\rightarrow$ T2I
& 0.10
& 16.44 & 0
& 9.22 & 82.94
& 1.447 & 19.098 & 8.442 & 8.382
& 51.56 \\

& 0.20
& 28.83 & 0
& 8.19 & 73.73
& 1.451 & 17.197 & 8.445 & 8.388
& 52.76 \\

& 0.50
& 65.80 & 0
& 5.12 & 46.08
& 1.454 & 22.850 & 8.505 & 8.451
& 65.13 \\

\midrule

Staged: Language $\rightarrow$ T2I$^\dagger$
& 0.33
& 68.03 & 0
& 10.24 & 92.16
& 1.445 & 18.809 & \textbf{8.428} & 8.371
& 49.22 \\

\bottomrule
\end{tabular}%
}

\end{table*}

%% file: tables/continual_pretraining_results.tex
\begin{table*}[!h]
\centering
\caption{\textbf{Continual pretraining from a language model.}
Qwen3-0.6B-CPT is initialized from Qwen3-0.6B-Base (pretrained on 36T text tokens), while Qwen3-0.6B is randomly initialized.
Both share the same Transformer architecture as \methodname{}-0.52B but use a larger vocabulary (216k vs.\ 80,640 for \methodname{}).
We additionally evaluate continual pretraining with a lower peak learning rate ($^*$: $1\times10^{-4}$ vs.\ $7.5\times10^{-4}$) to test whether aggressive weight updates diminish the benefits of pretrained language weights.
Since vocabularies differ, we report bits per byte (bpb) for both text and images.
Neither configuration improves T2I generation over random initialization.
All models are trained for one epoch on GPIC and evaluated with CFG 6.0.
LP denotes language pretraining tokens.
Token counts exclude padding.
}
\label{tab:qwen_init}
\vspace{4pt}
\small
\setlength{\tabcolsep}{5pt}

\resizebox{\textwidth}{!}{%
\begin{tabular}{@{}l r rc rc cccc c@{}}
\toprule
& &
\multicolumn{2}{c}{Text tokens (B)}
& \multicolumn{2}{c}{Image tokens (B)}
& \multicolumn{2}{c}{Text bpb $\downarrow$}
& \multicolumn{2}{c}{Image bpb ($\times10^{-2}$) $\downarrow$}
& \\
\cmidrule(lr){3-4}
\cmidrule(lr){5-6}
\cmidrule(lr){7-8}
\cmidrule(lr){9-10}
Model
& LP
& $\mathcal{T}\mid\varnothing$
& $\mathcal{T}\mid\mathcal{I}$
& $\mathcal{I}\mid\varnothing$
& $\mathcal{I}\mid\mathcal{T}$
& $\mathcal{T}\mid\varnothing$
& $\mathcal{T}\mid\mathcal{I}$
& $\mathcal{I}\mid\varnothing$
& $\mathcal{I}\mid\mathcal{T}$
& gFD $\downarrow$ \\
\midrule

\methodname{}-0.52B
& 0
& 4.05 & 0
& 10.24 & 92.16
& 0.433 & 5.565 & 6.333 & 6.286
& 48.02 \\

\midrule

Qwen3-0.6B
& 0
& 4.05 & 0
& 10.24 & 92.16
& 0.434 & 5.436 & \textbf{6.324} & \textbf{6.278}
& \textbf{46.61} \\

Qwen3-0.6B-CPT
& 36T
& 4.05 & 0
& 10.24 & 92.16
& 0.430 & \textbf{4.974} & 6.326 & 6.283
& 47.81 \\

Qwen3-0.6B-CPT~$^*$
& 36T
& 4.05 & 0
& 10.24 & 92.16
& \textbf{0.424} & 6.743 & 6.409 & 6.367
& 63.62 \\

\bottomrule
\end{tabular}%
}

\end{table*}

%% file: sections/experiment_results_effect_of_i2t_supervision.tex
\subsection{Results: Effect of I2T Supervision on T2I generation}

\noindent\textbf{I2T supervision as an additional task does not improve T2I generation.}
We investigate whether adding I2T supervision as a separate task through staged or mixed training improves T2I generation under a fixed pretraining budget (Table~\ref{tab:i2t_scheduling}).
At an I2T allocation of $0.1$, both schedules achieve gFD comparable to the T2I baseline (48.4 and 48.2 vs.\ 48.0).
Increasing the allocation to $0.5$ reduces conditional image tokens by half (92.2B $\rightarrow$ 46.1B), with staged training outperforming mixing (54.2 vs.\ 76.0 gFD), although neither improves upon the baseline.
Conversely, mixing substantially reduces conditional text NLL compared to staging (1.009 vs.\ 20.437), indicating better image captioning.
These results suggest that allocating part of the pretraining budget to I2T as a separate task does not benefit T2I generation, despite improving captioning.
\input{tables/t2i_effect}

\begin{figure}[]
\begin{adjustbox}{width=1\textwidth,center}
\begin{tabular}{c}
    \includegraphics[width=1\textwidth]
    {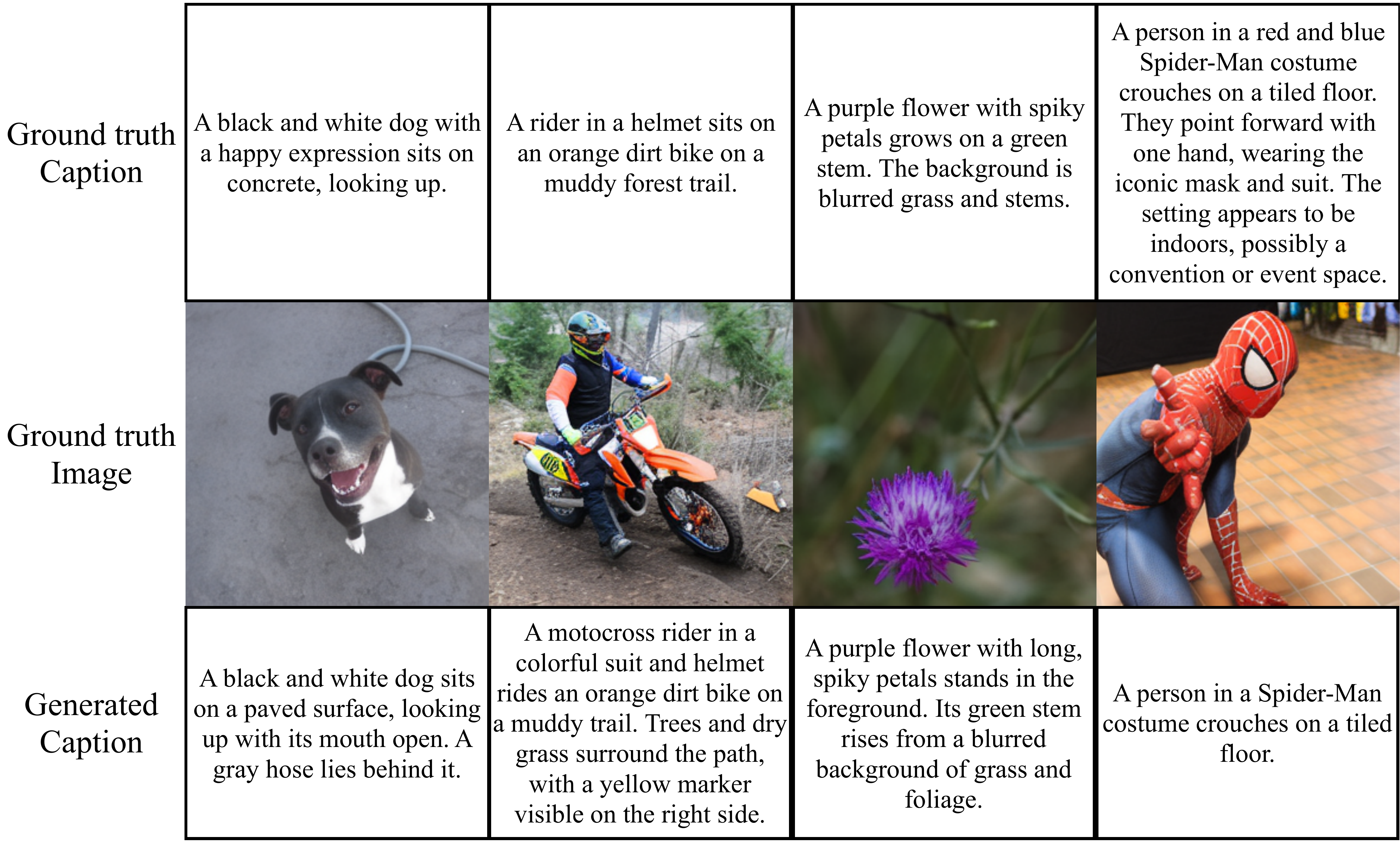} \\
\end{tabular}
\end{adjustbox}
\caption{\textbf{Image captioning examples from \methodname{}-4B.}
For each example, the ground-truth caption is shown above the input image, and the generated caption is shown below.
The model learns image captioning by replacing caption-dropped T2I examples with I2T supervision, using only 10\% of the pretraining budget for captioning.
}
\label{fig_main:captioning-examples}
\vspace{-0.2cm}
\end{figure}

\noindent\textbf{Replacing caption dropout with I2T supervision enables image captioning under the same pretraining budget.}
Rather than allocating a separate training budget to I2T, we replace caption-dropped T2I examples ($p_{\mathrm{drop}}=0.1$) with I2T examples ($p_{\mathrm{drop}}=0$, $\eta=0.1$).
Since I2T sequences share the $[\texttt{SOV}]\,x_1\cdots x_N\,[\texttt{EOV}]$ prefix, this preserves the conditional and unconditional image-token counts under the same pretraining budget.
Across all model sizes, conditional text NLL decreases by more than an order of magnitude (e.g., 18.414 $\rightarrow$ 1.102 at 0.52B), while gFD, GenEval, and DPG-Bench remain comparable to the baseline (Table~\ref{tab:joint_t2i_i2t}).
Caption alignment also increases with model size (0.090 $\rightarrow$ 0.559), with qualitative examples from \methodname{}-4.00B shown in Figure~\ref{fig_main:captioning-examples}.
Although caption alignment remains below that of GPIC ground truth 
captions, its consistent improvement with model size demonstrates the potential to learn additional multimodal capabilities while maintaining competitive T2I generation under the same pretraining budget.
\input{tables/caption_dropout_as_i2t_effect}

%% file: tables/t2i_effect.tex
\begin{table*}[!h]
\centering
\caption{\textbf{Effect of I2T supervision and scheduling on T2I generation for \methodname{}-0.52B.}
Under a fixed pretraining budget, staged and mixed training use identical token counts at each I2T allocation, differing only in scheduling.
At 10\%, both schedules preserve comparable T2I generation, while at 50\%, staged training achieves substantially lower gFD.
However, only mixing substantially reduces captioning NLL, namely $\mathrm{NLL}(\mathcal{T}\mid\mathcal{I})$.
NLL and gFD are evaluated on GPIC Test-50K; token counts exclude padding.
}
\label{tab:i2t_scheduling}
\vspace{4pt}
\small
\setlength{\tabcolsep}{5pt}

\resizebox{\textwidth}{!}{%
\begin{tabular}{@{}l c cc cc crcc c@{}}
\toprule
& &
\multicolumn{2}{c}{Text tokens (B)}
& \multicolumn{2}{c}{Image tokens (B)}
& \multicolumn{4}{c}{NLL $\downarrow$}
& \\
\cmidrule(lr){3-4}
\cmidrule(lr){5-6}
\cmidrule(lr){7-10}
Configuration
& $\beta$ (or $\eta$)
& $\mathcal{T}\mid\varnothing$
& $\mathcal{T}\mid\mathcal{I}$
& $\mathcal{I}\mid\varnothing$
& $\mathcal{I}\mid\mathcal{T}$

& $\mathcal{T}\mid\varnothing$
& $\mathcal{T}\mid\mathcal{I}$
& $\mathcal{I}\mid\varnothing$
& $\mathcal{I}\mid\mathcal{T}$
& gFD $\downarrow$ \\
\midrule

Baseline
& --
& 4.05 & 0
& 10.24 & 92.16
& \textbf{1.433} & 18.414 & 8.428 & 8.366
& \textbf{48.02} \\

\midrule

Staged: I2T $\rightarrow$ T2I
& $\beta=0.1$
& 3.65 & 0.45
& 19.46 & 82.94
& 1.440 & 18.415 & 8.426 & 8.366
& 48.35 \\

& $\beta=0.5$
& 2.03 & 2.25
& 56.32 & 46.08
& 1.435 & 20.437 & 8.421 & 8.370
& 54.22 \\

\midrule

Mixing: T2I + I2T
& $\eta=0.1$
& 3.65 & 0.45
& 19.46 & 82.94
& 1.443 & 1.125 & \textbf{8.414} & \textbf{8.362}
& 48.18 \\

& $\eta=0.5$
& 2.03 & 2.25
& 56.32 & 46.08
& 1.475 & \textbf{1.009} & 8.453 & 8.417
& 75.97 \\

\bottomrule
\end{tabular}%
}

\end{table*}

%% file: tables/caption_dropout_as_i2t_effect.tex
\begin{table*}[!bh]
\centering
\caption{\textbf{Replacing caption dropout with I2T supervision across model sizes.}
Replacing caption-dropped T2I examples with I2T preserves the pretraining budget and image-token counts, enabling image captioning while maintaining comparable T2I generation.
We report mean Qwen3-VL-Reranker-8B alignment scores over 50K generated image--caption pairs from GPIC Test-50K, using the image as the query and caption as the document.
GPIC Test-50k image-caption pairs obtain a mean alignment score of 0.891.
}
\label{tab:joint_t2i_i2t}
\vspace{4pt}
\small
\setlength{\tabcolsep}{5pt}

\resizebox{\textwidth}{!}{%
\begin{tabular}{@{}l cc cccc ccc c@{}}
\toprule
& & &
\multicolumn{4}{c}{NLL $\downarrow$}
& \multicolumn{3}{c}{T2I generation}
& Captioning \\
\cmidrule(lr){4-7}
\cmidrule(lr){8-10}
\cmidrule(l){11-11}
Size
& $\eta$
& $p_{\mathrm{drop}}$
& $\mathcal{T}\mid\varnothing$
& $\mathcal{T}\mid\mathcal{I}$
& $\mathcal{I}\mid\varnothing$
& $\mathcal{I}\mid\mathcal{T}$
& gFD $\downarrow$
& GenEval $\uparrow$
& DPG-Bench $\uparrow$
& Alignment $\uparrow$ \\
\midrule

\multirow{2}{*}{0.10B}
& --
& 0.1
& 1.601 & 19.066 & 8.774 & 8.727
& 144.16 & 27.31 & 66.36
& -- \\

& 0.1
& 0
& \textbf{1.598} & \textbf{1.410}
& \textbf{8.771} & \textbf{8.724}
& \textbf{144.14} & \textbf{27.66} & \textbf{68.89}
& 0.090 \\

\midrule

\multirow{2}{*}{0.25B}
& --
& 0.1
& 1.497 & 21.472 & 8.566 & 8.509
& 72.55 & 35.72 & 72.66
& -- \\

& 0.1
& 0
& \textbf{1.491} & \textbf{1.218}
& \textbf{8.559} & \textbf{8.503}
& \textbf{70.44} & \textbf{36.54} & \textbf{73.07}
& 0.207 \\

\midrule

\multirow{2}{*}{0.52B}
& --
& 0.1
& \textbf{1.433} & 18.414 & 8.428 & 8.366
& 48.02 & \textbf{44.91} & \textbf{76.84}
& -- \\

& 0.1
& 0
& 1.435 & \textbf{1.102}
& \textbf{8.419} & \textbf{8.358}
& \textbf{46.48} & 44.60 & 76.14
& 0.315 \\

\midrule

\multirow{2}{*}{1.6B}
& --
& 0.1
& \textbf{1.370} & 20.426 & 8.303 & 8.235
& \textbf{35.28} & 53.48 & 77.90
& -- \\

& 0.1
& 0
& 1.372 & \textbf{0.977}
& \textbf{8.292} & \textbf{8.228}
& 35.91 & \textbf{53.51} & \textbf{78.21}
& 0.467 \\

\midrule

\multirow{2}{*}{4B}
& --
& 0.1
& 1.338 & 16.738
& \textbf{8.193} & \textbf{8.122}
& \textbf{28.43} & \textbf{56.39} & 77.90
& -- \\

& 0.1
& 0
& \textbf{1.325} & \textbf{0.890}
& 8.196 & 8.126
& 29.30 & 55.33 & \textbf{78.61}
& 0.559 \\

\bottomrule
\end{tabular}%
}
\vspace{-0.4cm}
\end{table*}

%% file: sections/conclusion.tex
\newpage
\section{Discussion and Conclusion}

\begin{figure}[!h]
\begin{adjustbox}{width=1\textwidth,center}
\begin{tabular}{c}
    \includegraphics[width=1\textwidth]
    {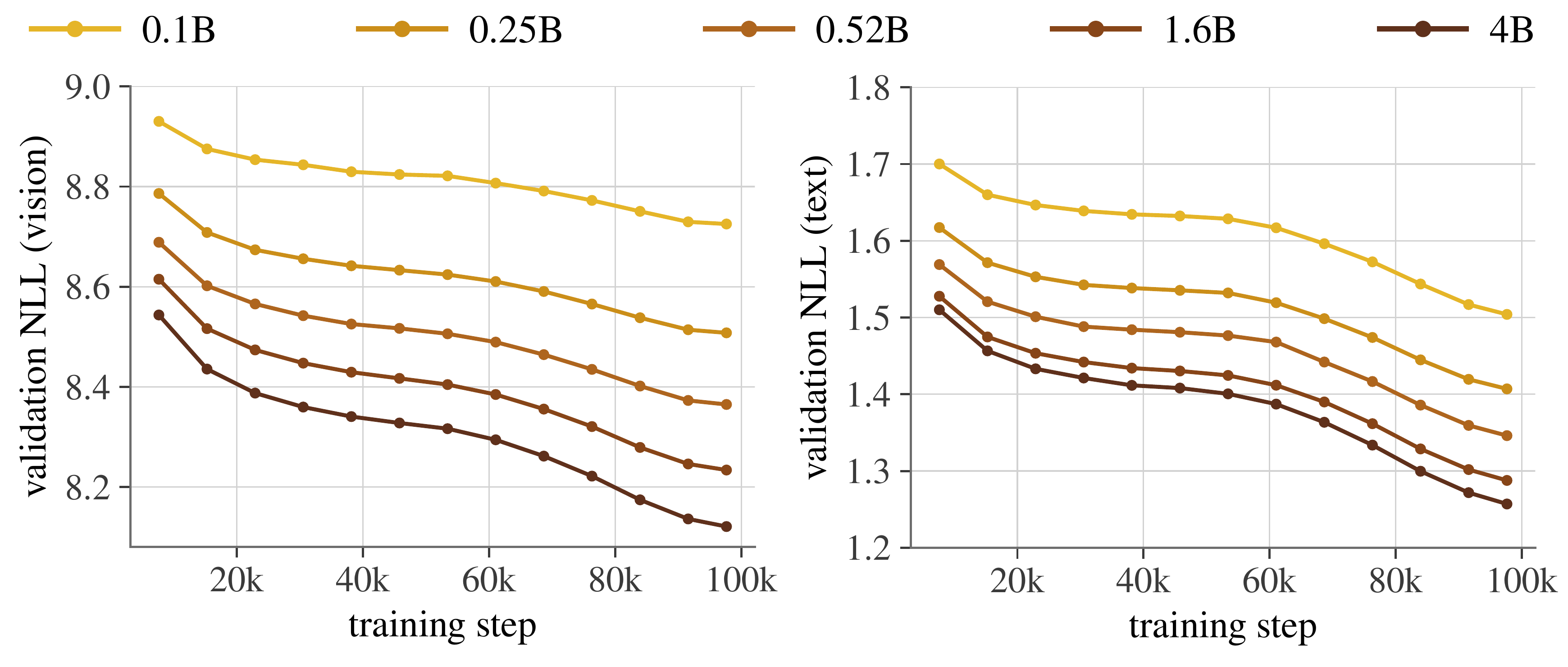} \\
\end{tabular}
\end{adjustbox}
\vspace{-0.4cm}
\caption{\textbf{Scaling behavior of \methodname{} across model sizes.}
Validation negative log-likelihood (NLL) for image tokens (left) and text tokens (right) throughout pretraining on GPIC.
As model size increases from 0.1B to 4B, both image and text NLL consistently decrease, with larger models achieving lower NLL throughout training.
All models are pretrained from scratch under the same training schedule.
}
\label{fig_main:scaling}
\vspace{-0.2cm}
\end{figure}

\noindent\textbf{Model scaling and classifier-free guidance.}
We examine how \methodname{} scales from 0.1B to 4B parameters.
Figure~\ref{fig_main:scaling} shows that both image and text validation NLL decrease consistently with model size, with larger models achieving lower NLL throughout pretraining.
This demonstrates improvements in modeling both modalities as model capacity increases.
We further examine how the optimal CFG scale varies with model size (Figure~\ref{fig:cfg_sweep}).
As model size increases from 0.1B to 4B, the guidance scale minimizing gFD on GPIC Validation-50K decreases from 7.5 to 4.5, while gFD improves from 142.7 to 28.4.
This suggests that larger models require less guidance to achieve their best generation quality, although CFG remains beneficial across all evaluated model sizes.

\noindent\textbf{Caption length and text conditioning.}
Since \methodname{} learns text conditioning directly from paired image--text data, we investigate how generation quality varies with the length of the caption.
We evaluate all five model sizes using short, medium, and long captions from GPIC Test-1M (Table~\ref{tab:caption_types}).
As model size increases from 0.1B to 4B, gFD decreases across all caption lengths.
Medium and long captions consistently achieve lower gFD than short captions.
These results suggest that more detailed text descriptions improve T2I generation quality, although longer captions do not consistently outperform medium-length captions.
\input{tables/caption_length_vs_gfd}

\noindent\textbf{Implications of language pretraining for T2I generation.}
Our results challenge the prevailing reliance on internet-scale language pretraining for T2I generation.
In late-fusion systems, pretrained language models provide text representations, but the image generator must learn to map these representations to images using paired data.
On GPIC Test-50K captions, FLAN-T5-XL achieves lower caption NLL than FLAN-T5-Base (2.76 vs.\ 3.24, averaged over three instruction prompts), with 13.5$\times$ more language-pretraining compute~\citep{chung2024scaling}.
Yet, using FLAN-T5-XL instead of FLAN-T5-Base only modestly improves LlamaGen gFD (66.7 $\rightarrow$ 63.3).
This result is consistent with Parti~\citep{yu2022scaling}, which reported marginal benefits from text-encoder pretraining.
For early-fusion models, text-only pretraining optimizes $p(\mathcal{T}\mid\varnothing)$, whereas T2I generation requires learning $p(\mathcal{I}\mid\mathcal{T})$ with image tokens produced by an independently trained tokenizer.
Strong text-only modeling therefore does not guarantee a useful initialization for image generation.
Consistent with this view, our randomly initialized Qwen3 model achieves lower vision validation NLL throughout pretraining (Appendix Figure~\ref{fig_supp:qwen3_cpt_nll}) and lower final gFD than its language-pretrained counterpart.
In contrast, \methodname{} learns text representations for image generation jointly from paired image--text data without language pretraining, although our experiments do not isolate the effect of joint training.
Since our study focuses on pretraining, we account for all pretraining FLOPs required to build each T2I system, including language-model, image-tokenizer, and image-generator pretraining.
We note that our study examines the role of language pretraining in late- and early-fusion systems, not the relative capabilities of autoregressive and diffusion-based image generators.
Beyond T2I, our findings motivate reconsidering the role of modality-specific pretraining in other multimodal settings, including biological data~\citep{omnii_health_preview_2026, data_mixing_2026, lee2026designing}.

\noindent\textbf{Limitations and future directions.}
While our experiments use GPIC, which is, to our knowledge, the largest high-quality, centrally hosted, publicly available dataset and benchmark for visual generation, extending these findings to other datasets and higher resolutions remains important.
Our controlled data composition and scheduling experiments use \methodname{}-0.52B, while the pretrained initialization comparison uses Qwen3-0.6B; extensions to larger model sizes and pretraining budgets would help establish the generality of our findings.
Although our late-fusion baselines include diffusion-based generators, our controlled interventions focus on autoregressive models.
Extending these experiments to Transfusion-style models~\citep{zhou2025transfusion} is a valuable direction.
Finally, text-to-video systems such as Wan~\citep{wan2025wan}, CogVideoX~\citep{yang2025cogvideox}, and Mochi~\citep{genmo2024mochi} also rely on large pretrained language encoders for video generation.
Investigating whether T2V models can learn text conditioning directly from paired video--text data, without language pretraining, is a promising direction toward efficient video generation systems.

\textbf{Conclusion.}
In this work, we study whether internet-scale language pretraining is necessary for strong T2I generation.
We introduced \methodname, a family of early-fusion multimodal autoregressive models trained entirely from scratch on paired image--text data (GPIC). 
Our models outperform strong late-fusion baselines while using significantly less total pretraining compute.
Our controlled experiments show that external text-only pretraining does not improve T2I generation under the evaluated settings.
Furthermore, replacing caption-dropped T2I examples with I2T supervision enables image captioning while maintaining competitive generation quality under the same pretraining budget.
Our results show that strong T2I systems can be trained from scratch without internet-scale language pretraining.
Code and models are available at 
\href{https://cs.stanford.edu/~keshik/halcyon}{\nolinkurl{cs.stanford.edu/~keshik/halcyon}}.

\section*{AI Use Statement}
In this work, we used generative AI tools to assist with implementing methods.
We did not use generative AI tools to propose or refine hypotheses, design or provide feedback on research methodology or experiments, support qualitative or thematic data analysis, or interpret results.
Generating synthetic datasets, developing theoretical models or conceptual frameworks, formulating mathematical claims, providing critical ingredients for proving mathematical claims, assisting in the writing of proofs, assisting with translation, and cleaning or reformatting datasets were not applicable to our work.
Additionally, we used generative AI tools to create or edit software code, search for and identify relevant literature and information, draft and edit parts of the manuscript, and improve readability.
We reviewed all AI-assisted work.
AI-generated or edited code was inspected and tested for correctness, and information and references suggested by AI tools were verified against the original sources.
We take responsibility for the final content of this work, including text, claims, code, and artifacts produced with the aid of generative AI.

\section*{Ethics Statement}
Our work studies text-to-image generative modeling, which can be used to generate realistic visual content and may therefore be misused. 
By providing a significantly lower-cost recipe for training strong T2I models, our work lowers the barrier to developing capable generative systems, including for malicious use. \methodname{} and the baseline models are pretrained on GPIC and may inherit biases or limitations present in its images and captions. Our work does not involve human-subject studies or the collection of a new dataset. 
We do not identify additional ethical risks specific to our method beyond those associated with visual generative models.

\section*{Reproducibility Statement}
We provide the model architecture, tokenization, training objective, data composition and scheduling, baseline modifications, evaluation protocol, and compute accounting in the main paper, with full architecture, optimization, and sampling details in the Appendix. The GPIC dataset and benchmark are publicly available. Our model checkpoints are publicly available.

%% file: tables/caption_length_vs_gfd.tex
\begin{figure}[!h]
\centering
\begin{minipage}[t]{0.45\textwidth}
  \vspace{0pt}
  \centering
  \includegraphics[width=0.99\linewidth]{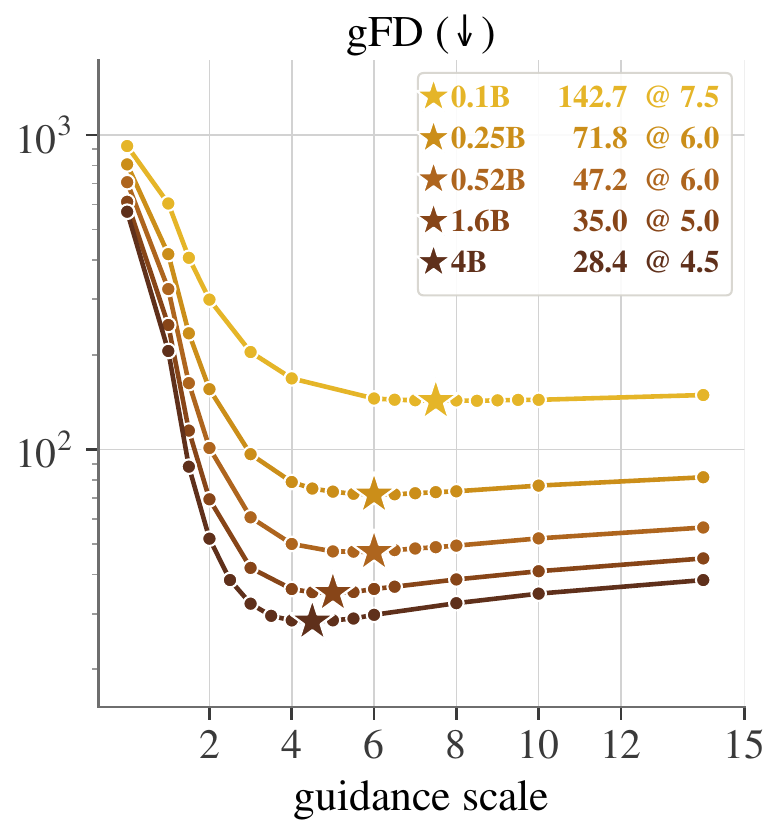}
  \vspace{-0.8cm}
  \caption{
  CFG sweep on GPIC validation. The optimal scale drops from 7.5 (0.1B) to 4.5 (4B) as gFD improves from 142.7 to 28.4; these scales are used on Test-50K.
  }
  \label{fig:cfg_sweep}
\end{minipage}\hfill
\begin{minipage}[t]{0.53\textwidth}
  \vspace{0pt}
  \centering
\captionof{table}{\textbf{Effect of caption length on T2I generation.}
We report gFD ($\downarrow$) for \methodname{} across model sizes using 50K examples with short, medium, and long captions from GPIC Test-1M.
Test-50K results are included for comparison.
Across all model sizes, medium and long captions achieve lower gFD than short captions, with comparable performance between medium and long captions from 0.52B onward.
Generation quality consistently improves with model size across all caption lengths.
}
  \label{tab:caption_types}
  \vspace{4pt}
  \small
  \begin{tabular}{l rrrr}
  \toprule
  Model & Test-50K & Short & Medium & Long \\
  \midrule
  0.1B  & 144.16 & 149.03 & 139.27 & 147.85 \\
  \midrule
  0.25B & 72.55  & 77.18  & 70.14  & 72.53 \\
   \midrule
  0.52B & 48.02  & 51.99  & 46.12  & 46.83 \\
   \midrule
  1.6B  & 35.28  & 38.59  & 34.21  & 34.33 \\
   \midrule
  4B    & 28.43  & 31.48  & 27.91  & 27.77 \\
  \bottomrule
  \end{tabular}
\end{minipage}
\end{figure}

%% file: sections/acknowledgements.tex
\section*{Acknowledgments}
We thank Radical Numerics for providing compute for this project. We thank Willie Neiswanger, Garyk Brixi, Tiange Xiang, Wenlong Huang, Mark Endo, Federico Berto, Zane Durante, Bailey Trang, Yanchen Wang, and Suzannah Wistreich for feedback on the manuscript. We also thank Suchir Agarwal for valuable discussions, the broader Radical Numerics research team for their feedback, and Alessandro Moro for infrastructure support.

%% file: sections_supp/model_configs.tex
\section{Training Details}
\label{sec:supp:model}

We provide model architecture and training details for all \methodname{}~models in Table~\ref{tab:configs}.

\begin{table}[!h]
\centering
\caption{\textbf{\methodname{}~configurations.}}
\label{tab:configs}
\small
\begin{tabular}{l|ccccc}
\toprule
 & \textbf{0.1B} & \textbf{0.25B} & \textbf{0.52B} & \textbf{1.6B} & \textbf{4B} \\
\midrule
\rowcolor{gray!15}\multicolumn{6}{l}{\textbf{architecture}} \\
backbone            & \multicolumn{5}{c}{Qwen3~\citep{yang2025qwen3}, decoder-only} \\
layers              & 14 & 21 & 28 & 28 & 36 \\
hidden dim          & 512 & 768 & 1024 & 2048 & 2560 \\
FFN dim             & 1536 & 2304 & 3072 & 6144 & 9728 \\
query / KV heads    & 8 / 4 & 12 / 6 & 16 / 8 & 16 / 8 & 32 / 8 \\
head dim            & \multicolumn{5}{c}{128} \\
non-embedding params& 55.1M & 185.8M & 440.4M & 1409.3M & 3634M \\
embedding params    & 41.3M & 61.9M & 82.6M & 165.2M & 206.4M \\
total params        & 96.4M & 247.7M & 523.0M & 1574.5M & 3840M \\
vocabulary          & \multicolumn{5}{c}{80{,}640 (16{,}384 text + 64{,}000 image + 256 special)} \\
tied embeddings     & \multicolumn{5}{c}{yes} \\
image size          & \multicolumn{5}{c}{256} \\
image tokens        & \multicolumn{5}{c}{1{,}024} \\
max caption tokens  & \multicolumn{5}{c}{254} \\
sequence length     & \multicolumn{5}{c}{1{,}280} \\
\midrule
\rowcolor{gray!15}\multicolumn{6}{l}{\textbf{training}} \\
training steps      & \multicolumn{5}{c}{97{,}632 (1 epoch on GPIC)} \\
global batch size   & \multicolumn{5}{c}{1{,}024} \\
warmup steps        & \multicolumn{5}{c}{1{,}000} \\
peak learning rate  & 1e-3 & 7.5e-4 & 7.5e-4 & 7.5e-4 & 7.5e-4 \\
lr schedule         & \multicolumn{5}{c}{linear warmup, constant to step 49{,}595, cooldown to $\text{peak}/10$} \\
optimizer           & \multicolumn{5}{c}{AdamW, $\beta_1,\beta_2 = 0.9, 0.95$} \\
weight decay        & \multicolumn{5}{c}{0.1} \\
gradient clipping   & \multicolumn{5}{c}{1.0} \\
init std            & \multicolumn{5}{c}{0.02} \\
image loss weight $\lambda_\mathrm{img}$ & \multicolumn{5}{c}{0.5} \\
caption dropout $p_\mathrm{drop}$ & \multicolumn{5}{c}{0.1} \\
precision           & \multicolumn{5}{c}{bf16} \\
\midrule
\rowcolor{gray!15}\multicolumn{6}{l}{\textbf{sampling}} \\
cfg  & 7.5 & 6.0 & 6.0 & 5.0 & 4.5 \\
\bottomrule
\end{tabular}
\end{table}

\begin{table}[t]
\centering
\caption{\textbf{JiT-T2I configurations.}}
\label{tab:jit_cfg}
\small
\begin{tabular}{lccc}
\toprule
 & JiT-T2I-XXL/16 & JiT-T2I-XXL/16++ & JiT-T2I-XXL/8++ \\
\midrule
\multicolumn{4}{l}{\textit{language model (frozen)}} \\
text encoder & \multicolumn{3}{c}{Qwen3-1.7B} \\
feature dim & \multicolumn{3}{c}{2048} \\
max caption tokens & \multicolumn{3}{c}{300} \\
\midrule
\multicolumn{4}{l}{\textit{image generator}} \\
backbone & \multicolumn{3}{c}{JiT-T2I (pixel space, no image tokenizer)} \\
blocks (image / text) & \multicolumn{3}{c}{16 / 4} \\
hidden dim & \multicolumn{3}{c}{1536} \\
attention heads & \multicolumn{3}{c}{24} \\
patch bottleneck dim & \multicolumn{3}{c}{256} \\
params & \multicolumn{3}{c}{1.10B} \\
patch size & 16 & 16 & 8 \\
image tokens & 256 & 256 & 1,024 \\
\midrule
\multicolumn{4}{l}{\textit{training}} \\
objective & \multicolumn{3}{c}{flow matching, linear schedule} \\
timestep sampling & \multicolumn{3}{c}{logit-normal ($\mu{=}{-}0.8$, $\sigma{=}0.8$), $t_\epsilon{=}0.05$} \\
LPIPS weight & -- & 0.1 & 0.1 \\
LPIPS timestep gate & -- & $t \ge 0.3$ & $t \ge 0.3$ \\
training steps & \multicolumn{3}{c}{390,600} \\
global batch size & \multicolumn{3}{c}{256} \\
peak learning rate & \multicolumn{3}{c}{1e-4} \\
lr schedule & \multicolumn{3}{c}{4,000 warmup, constant, 4,000 cooldown} \\
optimizer & \multicolumn{3}{c}{AdamW, $\beta_1,\beta_2 = 0.9, 0.95$} \\
weight decay & \multicolumn{3}{c}{0} \\
gradient clipping & \multicolumn{3}{c}{1.0} \\
EMA decay & \multicolumn{3}{c}{0.9999} \\
caption dropout & \multicolumn{3}{c}{0.1} \\
precision & \multicolumn{3}{c}{bf16} \\
\midrule
\multicolumn{4}{l}{\textit{sampling}} \\
sampler & \multicolumn{3}{c}{Euler (ODE), 50 steps, EMA weights} \\
CFG scale & 6.00 & 5.00 & 4.00 \\
\bottomrule
\end{tabular}
\end{table}

\begin{table}[t]
\centering
\caption{\textbf{LlamaGen configurations.}}
\label{tab:llamagen_cfg}
\small
\begin{tabular}{lccc}
\toprule
 & LlamaGen-XL & LlamaGen-XL/8 & LlamaGen-XL-T5-Base/8 \\
\midrule
\multicolumn{4}{l}{\textit{language model (frozen)}} \\
text encoder & FLAN-T5-XL & FLAN-T5-XL & FLAN-T5-Base \\
encoder params & 1.22B & 1.22B & 0.11B \\
feature dim & 2048 & 2048 & 768 \\
max caption tokens & \multicolumn{3}{c}{300} \\
\midrule
\multicolumn{4}{l}{\textit{image tokenizer}} \\
tokenizer & VQ-16 (pretrained) & \multicolumn{2}{c}{Ours (GPIC-trained)} \\
spatial downsampling & 16$\times$ & \multicolumn{2}{c}{8$\times$} \\
codebook size & 16,384 & \multicolumn{2}{c}{64,000} \\
image tokens & 256 & \multicolumn{2}{c}{1,024} \\
\midrule
\multicolumn{4}{l}{\textit{image generator}} \\
backbone & \multicolumn{3}{c}{GPT-XL (Llama-style, decoder-only)} \\
layers / hidden dim / heads & \multicolumn{3}{c}{36 / 1280 / 20} \\
FFN dim & \multicolumn{3}{c}{3584 (SwiGLU)} \\
params & 0.78B & 0.90B & 0.90B \\
sequence length & 556 & 1,324 & 1,324 \\
\midrule
\multicolumn{4}{l}{\textit{training}} \\
training steps & 390,600 & 390,600 & 390,600 \\
global batch size & \multicolumn{3}{c}{256} \\
peak learning rate & \multicolumn{3}{c}{1e-4} \\
lr schedule & \multicolumn{3}{c}{3,906 warmup, then constant} \\
optimizer & \multicolumn{3}{c}{AdamW, $\beta_1,\beta_2 = 0.9, 0.95$} \\
weight decay & \multicolumn{3}{c}{0.05} \\
gradient clipping & \multicolumn{3}{c}{1.0} \\
dropout / token dropout & \multicolumn{3}{c}{0.1 / 0.1} \\
caption dropout & \multicolumn{3}{c}{0.1} \\
precision & \multicolumn{3}{c}{bf16} \\
\midrule
\multicolumn{4}{l}{\textit{sampling}} \\
CFG scale & 
6.5 & 6.5 & 5.5 \\
temperature / top-$k$ / top-$p$ & 1.0 / 1000 / off & \multicolumn{2}{c}{1.0 / off / off} \\
\bottomrule
\end{tabular}
\end{table}

\begin{table}[t]
\centering
\caption{\textbf{Tokenizer training configurations.}}
\label{tab:tokenizer_configs}
\small
\begin{tabular}{ll}
\toprule
\multicolumn{2}{l}{\textbf{Image tokenizer}} \\
\midrule
architecture            & Cosmos~\citep{agarwal2025cosmos} encoder--decoder, FSQ~\citep{mentzer2024fsq} \\
decoder width           & 192 (128 in Cosmos) \\
spatial downsampling    & $8\times$ \\
codebook size           & 64{,}000 (FSQ levels $[8,8,8,5,5,5]$) \\
Params           & 136M \\
tokens per image        & 1{,}024 ($32 \times 32$, raster order) \\
losses (stage 1)        & reconstruction (1.0) + LPIPS (0.1) \\
losses (stage 2)        & + PatchGAN adversarial (5e-3) \\
training data           & GPIC train images, 1.5 epochs (1.0 stage 1 / 0.5 stage 2) \\
iterations              & 242{,}172 \\
batch size              & 640 \\
learning rate           & 1e-4, constant \\
optimizer               & Adam, $\beta_1, \beta_2 = 0.9, 0.999$ \\
weight decay            & 0.001 \\
gradient clipping       & 1.0 \\
EMA                     & yes \\
hardware                & $1 \times$ 8$\times$H100 node \\
\midrule
\multicolumn{2}{l}{\textbf{Text tokenizer}} \\
\midrule
algorithm               & byte-level BPE~\citep{sennrich2016bpe} \\
vocabulary size         & 16{,}384 \\
training data           & GPIC train captions (21.3~GB) + OpenWebText (11.9~GB) \\
caption length          & mean 44.5, p50 36, p90 64, p99 182 tokens \\
\bottomrule
\end{tabular}
\end{table}

%% file: tables/flops_vs_gfd.tex
\begin{table*}[t]
\centering
\caption{\textbf{Total pretraining compute of T2I systems.}
We report total pretraining FLOPs, decomposed into language, image-generator, and image-tokenizer compute, for all configurations shown in Fig.~\ref{fig_main:compute_vs_gfd}.
Vision compute includes both image-generator and image-tokenizer pretraining.
For late-fusion methods, language FLOPs include pretraining of the language model.
For \methodname{}, language FLOPs correspond to the text-token portion of joint multimodal pretraining, without external language pretraining.
All image generators are pretrained on GPIC for one epoch at $256\times256$ resolution.
Generation Fr\'echet distance (gFD) is evaluated on GPIC Test-50K; lower is better.}
\label{tab:pretraining_compute}
\small
\setlength{\tabcolsep}{4pt}
\adjustbox{max width=\textwidth}{%
\begin{tabular}{lrrrrrrr}
\toprule
& & \multicolumn{2}{c}{Language} & \multicolumn{3}{c}{Vision} & \\
\cmidrule(lr){3-4}
\cmidrule(lr){5-7}
Method
& Total FLOPs
& FLOPs
& (\%)
& Generator FLOPs
& Tokenizer FLOPs
& (\%)
& gFD $\downarrow$ \\
\midrule
LlamaGen-XL
& $9.32\times10^{21}$
& $9.06\times10^{21}$
& 97.17
& $1.19\times10^{20}$
& $1.45\times10^{20}$
& 2.83
& 147.01 \\

LlamaGen-XL/8
& $9.68\times10^{21}$
& $9.06\times10^{21}$
& 93.58
& $4.76\times10^{20}$
& $1.45\times10^{20}$
& 6.42
& 63.29 \\

LlamaGen-XL-T5-Base/8
& $1.29\times10^{21}$
& $6.69\times10^{20}$
& 51.87
& $4.76\times10^{20}$
& $1.45\times10^{20}$
& 48.13
& 66.69 \\

\midrule

JiT-T2I-XXL/16
& $3.67\times10^{23}$
& $3.67\times10^{23}$
& 99.95
& $1.69\times10^{20}$
& $0$
& 0.05
& 67.92 \\

JiT-T2I-XXL/16++
& $3.67\times10^{23}$
& $3.67\times10^{23}$
& 99.95
& $1.69\times10^{20}$
& $0$
& 0.05
& 48.86 \\

JiT-T2I-XXL/8++
& $3.68\times10^{23}$
& $3.67\times10^{23}$
& 99.82
& $6.76\times10^{20}$
& $0$
& 0.18
& 40.97 \\

\midrule

\methodname{}-0.1B
& $2.05\times10^{20}$
& $1.47\times10^{19}$
& 7.15
& $5.93\times10^{19}$
& $1.31\times10^{20}$
& 92.85
& 144.16 \\

\methodname{}-0.25B
& $3.21\times10^{20}$
& $3.78\times10^{19}$
& 11.74
& $1.53\times10^{20}$
& $1.31\times10^{20}$
& 88.26
& 72.55 \\

\methodname{}-0.52B
& $5.33\times10^{20}$
& $7.97\times10^{19}$
& 14.96
& $3.22\times10^{20}$
& $1.31\times10^{20}$
& 85.04
& 48.02 \\

\methodname{}-1.6B
& $1.34\times10^{21}$
& $2.40\times10^{20}$
& 17.90
& $9.69\times10^{20}$
& $1.31\times10^{20}$
& 82.10
& 35.28 \\

\methodname{}-4B
& $3.08\times10^{21}$
& $5.85\times10^{20}$
& 19.00
& $2.36\times10^{21}$
& $1.31\times10^{20}$
& 81.00
& 28.43 \\

\bottomrule
\end{tabular}%
}
\end{table*}

%% file: sections_supp/sampling_details.tex
\section{Sampling Details}
\label{app:sampling}

All models generate one $256\times256$ image per caption using the GPIC Test-50K set. For \methodname{},~we sample 1,024 image tokens autoregressively after the caption prefix and \texttt{[SOV]}, using temperature 1.0 with top-k and top-p disabled and a fixed seed. CFG is applied only to image-token logits, $\ell_{\text{cfg}}=\ell_u+\gamma(\ell_c-\ell_u)$, where the unconditional branch uses the \texttt{[SOV]} prefix without the caption. For each model size, we select $\gamma$ to minimize gFD on the GPIC validation set (Figure~\ref{fig:cfg_sweep}).
\textbf{Baselines.} For LlamaGen and JiT-T2I, we use the official sampling code and settings and identify the optimal CFG setting using the GPIC validation set. 

\input{tables/cfg_list_supp}

%% file: tables/cfg_list_supp.tex
\begin{table*}[!h]
\centering
\caption{\textbf{Classifier-free guidance (CFG) scales for T2I evaluation.}
We report the CFG scale selected for each model based on the lowest gFD on GPIC Validation-50K. Sweeps are shown in Figures \ref{fig:cfg_sweep} and \ref{fig_supp:cfg_sweep_llamagen_jit}.
}
\label{tab:t2i_cfg}
\small
\setlength{\tabcolsep}{3.2pt}

\adjustbox{max width=0.5\textwidth}{%
\begin{tabular}{@{}lccc@{}}
\toprule
Model
& $p_{\mathrm{drop}}$
& $\eta$
& CFG Scale \\
\midrule

LlamaGen-XL
& 0.1 & -- & 6.5 \\

LlamaGen-XL/8
& 0.1 & -- & 6.5 \\

LlamaGen-XL-T5-Base/8
& 0.1 & -- & 5.5 \\

\midrule

JiT-T2I-XXL/16
& 0.1 & -- & 6.0 \\

JiT-T2I-XXL/16 ++
& 0.1 & -- & 5.0 \\

JiT-T2I-XXL/8 ++
& 0.1 & -- & 4.0 \\

\midrule

& 0.1 & -- & 7.5 \\
\multirow{-2}{*}{\methodname{}-0.10B}
& 0.0 & 0.1 & 7.5 \\

& 0.1 & -- & 6.0 \\
\multirow{-2}{*}{\methodname{}-0.25B}
& 0.0 & 0.1 & 6.0 \\

& 0.1 & -- & 6.0 \\
\multirow{-2}{*}{\methodname{}-0.52B}
& 0.0 & 0.1 & 6.0 \\

& 0.1 & -- & 5.0 \\
\multirow{-2}{*}{\methodname{}-1.6B}
& 0.0 & 0.1 & 5.2 \\

& 0.1 & -- & 4.5 \\
\multirow{-2}{*}{\methodname{}-4B}
& 0.0 & 0.1 & 4.6 \\

\bottomrule
\end{tabular}
}
\end{table*}

\begin{figure}[t]
  \centering
  \includegraphics[width=0.45\linewidth]{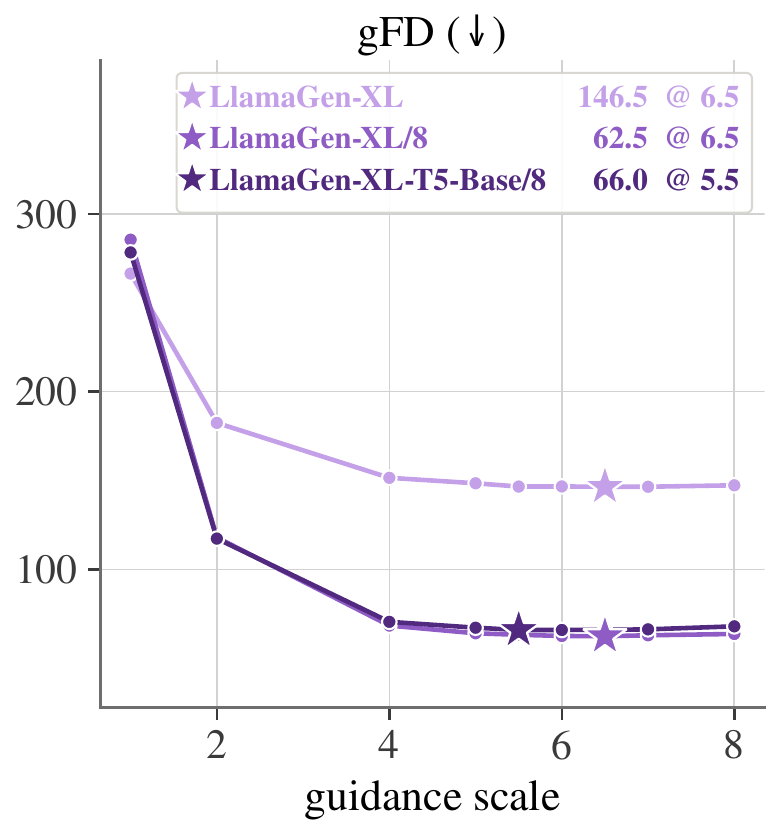}\hfill
  \includegraphics[width=0.45\linewidth]{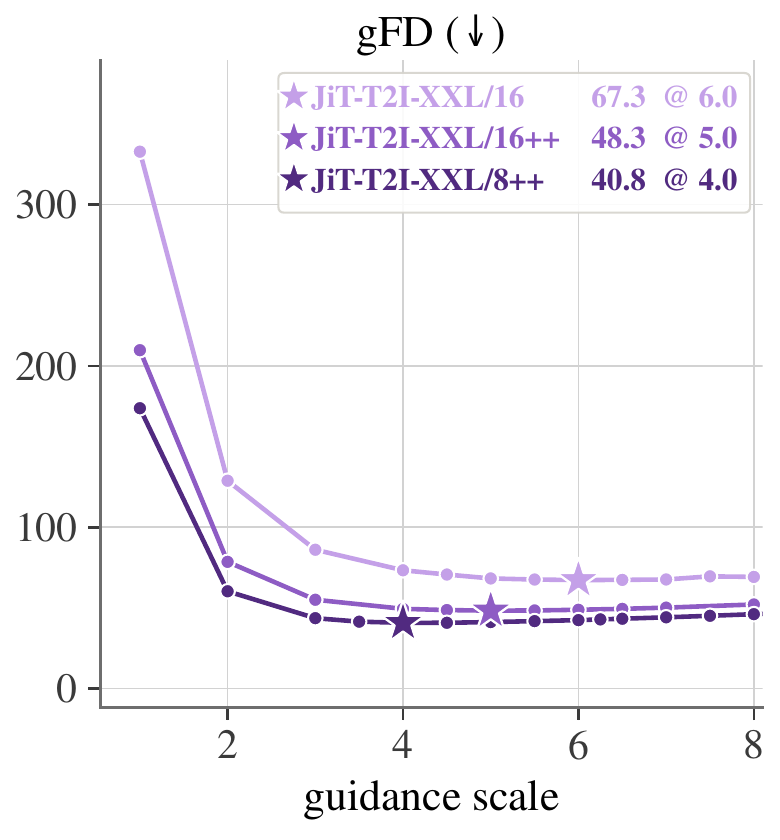}
  \caption{\textbf{Classifier-free guidance (CFG) sweeps for late-fusion T2I baselines.}
    We evaluate LlamaGen variants (left) and JiT-T2I variants (right) across CFG scales on GPIC Validation-50K.
    Stars indicate the scale achieving the lowest gFD for each configuration, with the corresponding gFD and CFG scale reported in the figure.
    The selected CFG scales are used for all subsequent evaluations on GPIC Test-50K, GenEval, and DPG-Bench.
    }
  \label{fig_supp:cfg_sweep_llamagen_jit}
\end{figure}

\input{tables/inference_cost}

\begin{figure}[t]
  \centering
  \includegraphics[width=0.95\linewidth]{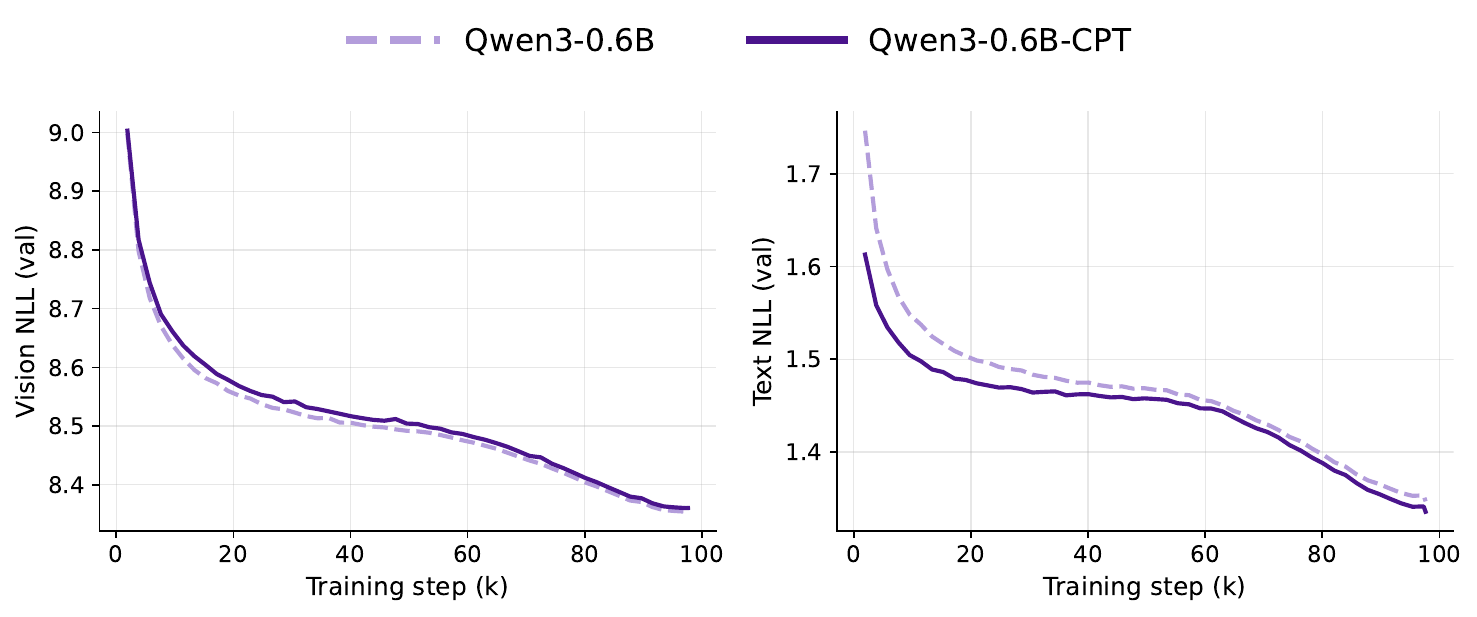}\hfill
  \caption{\textbf{Effect of language pretraining on validation NLL.}
Validation NLL for image tokens conditioned on text ($\mathcal{I}\mid\mathcal{T}$, left) and text tokens without conditioning ($\mathcal{T}\mid\varnothing$, right) during training of Qwen3-0.6B with random vs. language-pretrained initialization.
Random initialization consistently achieves lower image NLL, while language-pretrained initialization achieves lower text NLL.
}
\label{fig:qwen_init_nll}
  \label{fig_supp:qwen3_cpt_nll}
\end{figure}

%% file: tables/inference_cost.tex
\begin{table}[]
\centering
\caption{\textbf{Inference cost of Halcyon.}
Wall-clock time and throughput for generating 50K images (GPIC Test-50K set) using 8xH100 GPUs with vLLM, averaged over three runs.
We use temperature 1.0 without top-$k$/top-$p$ filtering and each model's CFG scale (two forward passes per image token).
Image decoding takes 198.0 $\pm$ 5.7 seconds (averaged over 3 runs) for all model sizes and is excluded from the reported results.}
\label{tab:inference_time}
\small
\begin{tabular}{lcc}
\toprule
Params & Wall time (s) $\downarrow$ & Throughput (img/s) $\uparrow$ \\
\midrule
0.1B  & 395.3 & 126.5 \\
0.25B & 507.4 &  98.5 \\
0.52B & 662.8 &  75.4 \\
1.6B  & 712.2 &  70.2 \\
4B    & 935.5 &  53.4 \\
\bottomrule
\end{tabular}
\end{table}

%% file: sections_supp/additional_results.tex
\section{Additional Results}
\label{sec_supp:additional_results}

\subsection{Per-Category Results on GenEval and DPG-Bench}
\label{sec:supp_geneval_breakdown}

Tabs.~\ref{tab:supp_geneval_full} and~\ref{tab:supp_dpg_full} report per-category GenEval and DPG-Bench results for all models in Tab.~\ref{tab:t2i_geneval_dpg}. GenEval overall scores are reported as mean $\pm$ std. We additionally report \methodname{} trained with I2T supervision ($p_\text{drop}{=}0.0$, $\eta{=}0.1$) at every scale. 
On GenEval, I2T supervision matches the standard recipe at all model sizes, with differences in overall scores within one standard deviation. On DPG-Bench, it yields slightly higher overall scores for four of the five model sizes.

\input{tables/geneval_breakdown}

%% file: tables/geneval_breakdown.tex
\begin{table}[t]
\centering
\caption{\textbf{GenEval results by category.} Overall is reported as mean $\pm$ std. ``+ I2T 10\%'' denotes the I2T supervision recipe ($p_\text{drop}{=}0.0$, $\eta{=}0.1$).}
\label{tab:supp_geneval_full}
\small
\setlength{\tabcolsep}{3.2pt}
\adjustbox{max width=\linewidth}{%
\begin{tabular}{l ccccccc}
\toprule
Model
& Overall $\uparrow$
& Single Obj.
& Two Obj.
& Counting
& Colors
& Position
& Color Attr. \\
\midrule

LlamaGen-XL
& 30.03\,{\scriptsize$\pm$1.10} & 78.44 & 19.95 & 20.31 & 53.46 & 4.25 & 3.75 \\

LlamaGen-XL/8
& 30.48\,{\scriptsize$\pm$1.74} & 75.94 & 23.74 & 15.94 & 53.99 & 9.50 & 3.75 \\

LlamaGen-XL-T5-Base/8
& 28.48\,{\scriptsize$\pm$1.19} & 78.12 & 18.18 & 15.94 & 48.40 & 6.75 & 3.50 \\

\midrule

JiT-T2I-XXL/16
& 36.27\,{\scriptsize$\pm$0.93} & 75.62 & 32.32 & 22.50 & 56.65 & 18.50 & 12.00 \\

JiT-T2I-XXL/16 ++
& 41.79\,{\scriptsize$\pm$2.48} & 85.00 & 41.41 & 21.88 & 61.97 & 23.25 & 17.25 \\

JiT-T2I-XXL/8 ++
& 43.17\,{\scriptsize$\pm$1.29} & 83.75 & 44.95 & 25.94 & 60.11 & 27.25 & 17.00 \\

\midrule

\rowcolor{mapit!30}
\methodname{}-0.10B
& 27.31\,{\scriptsize$\pm$0.61} & 70.94 & 26.52 & 8.12 & 46.54 & 7.25 & 4.50 \\

\rowcolor{mapit!30}
\quad + I2T 10\%
& 27.66\,{\scriptsize$\pm$2.32} & 76.88 & 20.71 & 9.06 & 47.07 & 6.50 & 5.75 \\

\rowcolor{mapit!45}
\methodname{}-0.25B
& 35.72\,{\scriptsize$\pm$1.68} & 78.44 & 35.35 & 18.75 & 55.05 & 13.50 & 13.25 \\

\rowcolor{mapit!45}
\quad + I2T 10\%
& 36.54\,{\scriptsize$\pm$2.02} & 78.12 & 39.39 & 15.31 & 56.91 & 16.25 & 13.25 \\

\rowcolor{mapit!60}
\methodname{}-0.52B
& 44.91\,{\scriptsize$\pm$2.29} & 83.75 & 50.00 & 26.56 & 61.17 & 29.75 & 18.25 \\

\rowcolor{mapit!60}
\quad + I2T 10\%
& 44.60\,{\scriptsize$\pm$1.25} & 81.88 & 51.52 & 22.81 & 60.64 & 29.50 & 21.25 \\

\rowcolor{mapit!80}
\methodname{}-1.60B
& 53.48\,{\scriptsize$\pm$1.01} & 88.75 & 67.68 & 28.12 & 63.56 & 42.00 & 30.75 \\

\rowcolor{mapit!80}
\quad + I2T 10\%
& 53.51\,{\scriptsize$\pm$3.01} & 85.94 & 63.13 & 28.75 & 66.76 & 47.75 & 28.75 \\

\rowcolor{mapit}
\methodname{}-4.00B
& 56.39\,{\scriptsize$\pm$1.33} & 92.81 & 68.18 & 29.06 & 67.02 & 51.00 & 30.25 \\

\rowcolor{mapit}
\quad + I2T 10\%
& 55.33\,{\scriptsize$\pm$0.55} & 87.81 & 62.88 & 28.75 & 71.28 & 44.00 & 37.25 \\

\bottomrule
\end{tabular}
}
\end{table}

\begin{table}[t]
\centering
\caption{\textbf{DPG-Bench results by category.} ``+ I2T 10\%'' denotes the I2T supervision recipe ($p_\text{drop}{=}0.0$, $\eta{=}0.1$).}
\label{tab:supp_dpg_full}
\small
\setlength{\tabcolsep}{3.2pt}
\adjustbox{max width=\linewidth}{%
\begin{tabular}{l cccccc}
\toprule
Model
& Overall $\uparrow$
& Global
& Entity
& Attribute
& Relation
& Other \\
\midrule

LlamaGen-XL
& 65.40 & 85.85 & 80.48 & 79.19 & 79.92 & 77.97 \\

LlamaGen-XL/8
& 67.08 & 79.09 & 80.53 & 79.24 & 86.21 & 79.30 \\

LlamaGen-XL-T5-Base/8
& 64.67 & 77.34 & 78.71 & 80.15 & 80.88 & 75.41 \\

\midrule

JiT-T2I-XXL/16
& 72.69 & 81.86 & 83.69 & 85.15 & 88.36 & 83.28 \\

JiT-T2I-XXL/16 ++
& 76.58 & 84.88 & 87.39 & 86.46 & 86.70 & 87.65 \\

JiT-T2I-XXL/8 ++
& 76.85 & 86.14 & 84.18 & 84.97 & 90.21 & 89.93 \\

\midrule

\rowcolor{mapit!30}
\methodname{}-0.10B
& 66.36 & 78.33 & 79.02 & 82.02 & 81.20 & 79.56 \\

\rowcolor{mapit!30}
\quad + I2T 10\%
& 68.89 & 75.37 & 79.85 & 83.88 & 82.36 & 82.06 \\

\rowcolor{mapit!45}
\methodname{}-0.25B
& 72.66 & 82.83 & 83.96 & 84.75 & 84.39 & 84.37 \\

\rowcolor{mapit!45}
\quad + I2T 10\%
& 73.07 & 81.75 & 81.59 & 86.03 & 85.76 & 85.58 \\

\rowcolor{mapit!60}
\methodname{}-0.52B
& 76.84 & 77.51 & 85.84 & 86.51 & 88.72 & 85.33 \\

\rowcolor{mapit!60}
\quad + I2T 10\%
& 76.14 & 83.59 & 85.61 & 86.62 & 86.35 & 87.14 \\

\rowcolor{mapit!80}
\methodname{}-1.60B
& 77.90 & 85.80 & 86.83 & 86.62 & 87.86 & 87.19 \\

\rowcolor{mapit!80}
\quad + I2T 10\%
& 78.21 & 91.66 & 84.76 & 87.98 & 86.36 & 84.36 \\

\rowcolor{mapit}
\methodname{}-4.00B
& 77.90 & 85.63 & 83.94 & 86.03 & 89.10 & 87.09 \\

\rowcolor{mapit}
\quad + I2T 10\%
& 78.61 & 89.02 & 86.98 & 86.20 & 90.16 & 84.50 \\

\bottomrule
\end{tabular}
}
\end{table}

%% file: sections_supp/training_seed_variance.tex
\subsection{Standard Deviation (Training Seeds)}
\label{sec:supp_seed_variance}

To estimate variance due to pretraining, we train \methodname{}-0.10B with I2T supervision ($p_\text{drop}{=}0.0$, $\eta{=}0.1$) three times with different training seeds, keeping all other settings fixed. We evaluate each run with CFG $7.5$. Tab.~\ref{tab:supp_seed_variance} reports per-seed results along with the mean $\pm$ std. The std of gFD across seeds is $5.54$, which gives a reference scale for interpreting gFD differences between small models. The sample-quality metrics (precision, recall, density, coverage) remain stable across seeds.

\begin{table}[t]
\centering
\caption{\textbf{Standard deviation of training runs.}
\methodname{}-0.10B with 10\% I2T supervision, pretrained using three random seeds and evaluated with CFG $7.5$.
The last row reports mean $\pm$ standard deviation (sample std, $n{=}3$).}
\label{tab:supp_seed_variance}
\small
\setlength{\tabcolsep}{3.2pt}
\adjustbox{max width=\linewidth}{%
\begin{tabular}{l cccccc}
\toprule
Seed
& gFD $\downarrow$
& Precision $\uparrow$
& Recall $\uparrow$
& Density $\to 1$
& Coverage $\uparrow$
& MMD $\downarrow$ \\
\midrule
1234 & 144.14 & 0.926 & 0.631 & 1.057 & 0.841 & 0.132 \\
2026 & 154.63 & 0.926 & 0.606 & 1.070 & 0.848 & 0.146 \\
7777 & 152.50 & 0.927 & 0.616 & 1.031 & 0.850 & 0.143 \\
\midrule
\rowcolor{mapit!30}
Mean $\pm$ std
& 150.42\,{\scriptsize$\pm$5.54}
& 0.926\,{\scriptsize$\pm$0.001}
& 0.618\,{\scriptsize$\pm$0.013}
& 1.053\,{\scriptsize$\pm$0.020}
& 0.846\,{\scriptsize$\pm$0.005}
& 0.140\,{\scriptsize$\pm$0.007} \\
\bottomrule
\end{tabular}
}
\end{table}

%% file: sections_supp/sampling_seed_variance.tex
\subsection{Standard Deviation (Sampling Seeds)}
\label{sec:supp_sampling_seed}

To measure how much evaluation noise comes from sampling, we generate the $50$k evaluation images for each \methodname{} model three times with different sampling seeds, keeping the checkpoint and CFG scale fixed. Tab.~\ref{tab:supp_sampling_seed_variance} shows that sampling variance is small at every scale: the std of gFD is at most $0.14$.

\begin{table}[t]
\centering
\caption{\textbf{Standard deviation of sampling runs.}
Each \methodname{} model is evaluated using a fixed checkpoint and three random sampling seeds (50K images per seed).
We report mean $\pm$ standard deviation ($n{=}3$).}
\label{tab:supp_sampling_seed_variance}
\small
\setlength{\tabcolsep}{3.2pt}
\adjustbox{max width=\linewidth}{%
\begin{tabular}{l c c cccccc}
\toprule
Model & CFG & Seed
& gFD $\downarrow$
& Precision $\uparrow$
& Recall $\uparrow$
& Density $\to 1$
& Coverage $\uparrow$
& MMD $\downarrow$ \\
\midrule

\methodname{}-0.10B & 7.5 & 1234 & 144.16 & 0.924 & 0.636 & 1.029 & 0.853 & 0.131 \\
 & & 2026 & 143.91 & 0.927 & 0.638 & 1.032 & 0.863 & 0.130 \\
 & & 7777 & 144.13 & 0.925 & 0.644 & 1.042 & 0.853 & 0.131 \\
\rowcolor{mapit!30}
 & & Avg.\ $\pm$ std
& 144.07\,{\scriptsize$\pm$0.14} & 0.925\,{\scriptsize$\pm$0.002} & 0.639\,{\scriptsize$\pm$0.004} & 1.034\,{\scriptsize$\pm$0.007} & 0.856\,{\scriptsize$\pm$0.006} & 0.131\,{\scriptsize$\pm$0.001} \\
\midrule

\methodname{}-0.25B & 6.0 & 1234 & 72.55 & 0.946 & 0.796 & 1.050 & 0.922 & 0.050 \\
 & & 2026 & 72.35 & 0.947 & 0.804 & 1.053 & 0.922 & 0.049 \\
 & & 7777 & 72.29 & 0.949 & 0.808 & 1.065 & 0.923 & 0.050 \\
\rowcolor{mapit!45}
 & & Avg.\ $\pm$ std
& 72.40\,{\scriptsize$\pm$0.14} & 0.947\,{\scriptsize$\pm$0.002} & 0.803\,{\scriptsize$\pm$0.006} & 1.056\,{\scriptsize$\pm$0.008} & 0.922\,{\scriptsize$\pm$0.001} & 0.050\,{\scriptsize$\pm$0.001} \\
\midrule

\methodname{}-0.52B & 6.0 & 1234 & 48.02 & 0.953 & 0.858 & 1.064 & 0.938 & 0.027 \\
 & & 2026 & 48.03 & 0.954 & 0.861 & 1.067 & 0.944 & 0.028 \\
 & & 7777 & 47.90 & 0.953 & 0.859 & 1.073 & 0.944 & 0.028 \\
\rowcolor{mapit!60}
 & & Avg.\ $\pm$ std
& 47.98\,{\scriptsize$\pm$0.08} & 0.953\,{\scriptsize$\pm$0.001} & 0.859\,{\scriptsize$\pm$0.002} & 1.068\,{\scriptsize$\pm$0.005} & 0.942\,{\scriptsize$\pm$0.003} & 0.028\,{\scriptsize$\pm$0.001} \\
\midrule

\methodname{}-1.60B & 5.0 & 1234 & 35.28 & 0.955 & 0.891 & 1.071 & 0.955 & 0.019 \\
 & & 2026 & 35.46 & 0.954 & 0.891 & 1.050 & 0.952 & 0.019 \\
 & & 7777 & 35.46 & 0.953 & 0.891 & 1.057 & 0.951 & 0.019 \\
\rowcolor{mapit!80}
 & & Avg.\ $\pm$ std
& 35.40\,{\scriptsize$\pm$0.11} & 0.954\,{\scriptsize$\pm$0.001} & 0.891\,{\scriptsize$\pm$0.000} & 1.059\,{\scriptsize$\pm$0.011} & 0.953\,{\scriptsize$\pm$0.002} & 0.019\,{\scriptsize$\pm$0.000} \\
\midrule

\methodname{}-4.00B & 4.5 & 1234 & 28.43 & 0.951 & 0.905 & 1.034 & 0.960 & 0.014 \\
 & & 2026 & 28.56 & 0.955 & 0.910 & 1.038 & 0.960 & 0.014 \\
 & & 7777 & 28.63 & 0.956 & 0.905 & 1.058 & 0.966 & 0.014 \\
\rowcolor{mapit}
 & & Avg.\ $\pm$ std
& 28.54\,{\scriptsize$\pm$0.10} & 0.954\,{\scriptsize$\pm$0.003} & 0.907\,{\scriptsize$\pm$0.003} & 1.043\,{\scriptsize$\pm$0.013} & 0.962\,{\scriptsize$\pm$0.003} & 0.014\,{\scriptsize$\pm$0.000} \\

\bottomrule
\end{tabular}
}
\end{table}

%% file: sections_supp/inference_time.tex
\subsection{Inference Cost}
\label{sec:inference_cost}

We measure inference cost on the GPIC 50k test set. 
For each model, we generate all 50k images on a single H100 node of 8 GPUs with vLLM~\citep{kwon2023efficient}, splitting prompts evenly across GPUs. We report wall-clock time averaged over 3 runs. 
Sampling settings match our evaluation settings.
We report results in Table~\ref{tab:inference_time}.


%% file: sections_supp/visualize.tex
\section{Samples}
\label{sec:supp:visualize}

\begin{figure}
    \centering
    \includegraphics[width=\linewidth]{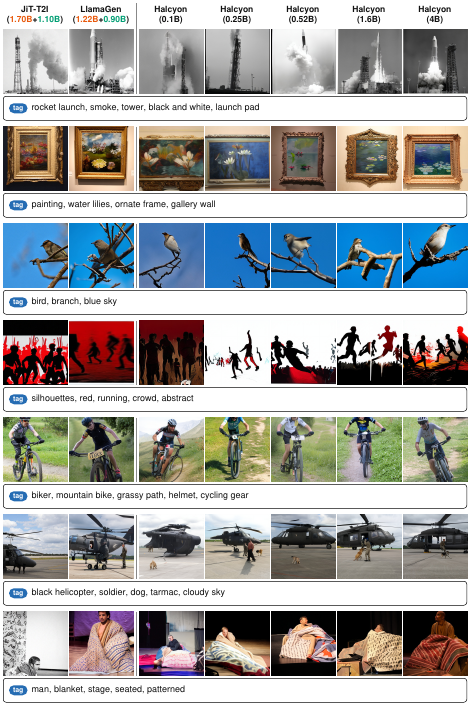}
    \vspace{-20pt}
    \caption{\textbf{Comparison of generated images with ``tag" captions.}
    From left to right: JiT-T2I-XXL/8++, LlamaGen-XL/8, and \methodname{} from 0.1B to 4B.
    For the baselines, we report the sizes of the {\color{orange}\bf language model} and {\color{darkgreen}\bf image generator} separately.
    Image quality improves as \methodname{} scales up. All prompts are from GPIC Test-50K.
    }
    \label{fig:supp:collage_tag}
\end{figure}

\begin{figure}
    \centering
    \includegraphics[width=\linewidth]{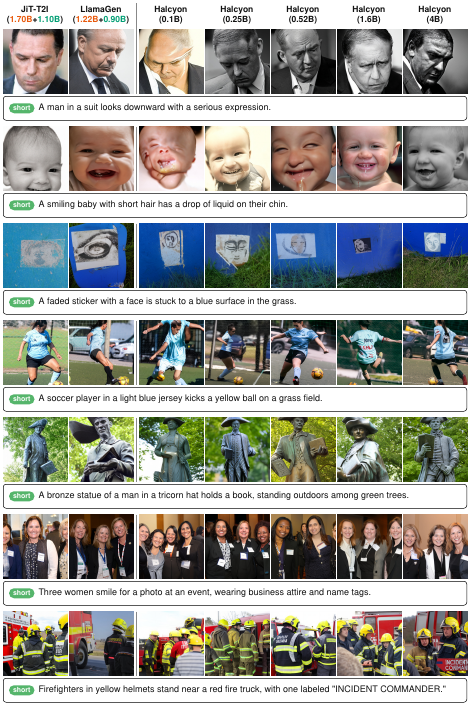}
    \vspace{-20pt}
    \caption{\textbf{Comparison of generated images with ``short" captions.}
    From left to right: JiT-T2I-XXL/8++, LlamaGen-XL/8, and \methodname{} from 0.1B to 4B.
    For the baselines, we report the sizes of the {\color{orange}\bf language model} and {\color{darkgreen}\bf image generator} separately.
    Image quality improves as \methodname{} scales up.
    All prompts are from GPIC Test-50K.
    }
    \label{fig:supp:collage_short}
\end{figure}

\begin{figure}
    \centering
    \includegraphics[width=\linewidth]{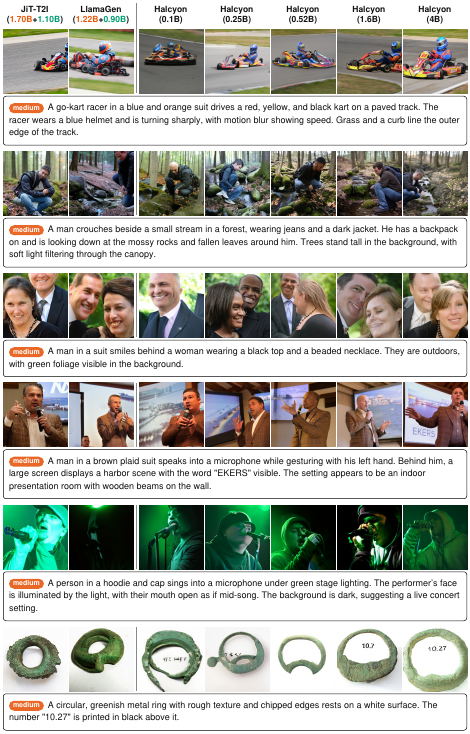}
    \vspace{-20pt}
    \caption{\textbf{Comparison of generated images with ``medium" captions.}
    From left to right: JiT-T2I-XXL/8++, LlamaGen-XL/8, and \methodname{} from 0.1B to 4B.
    For the baselines, we report the sizes of the {\color{orange}\bf language model} and {\color{darkgreen}\bf image generator} separately.
    Image quality improves as \methodname{} scales up.
    All prompts are from GPIC Test-50K.
    }
    \label{fig:supp:collage_medium}
\end{figure}

\begin{figure}
    \centering
    \includegraphics[width=\linewidth]{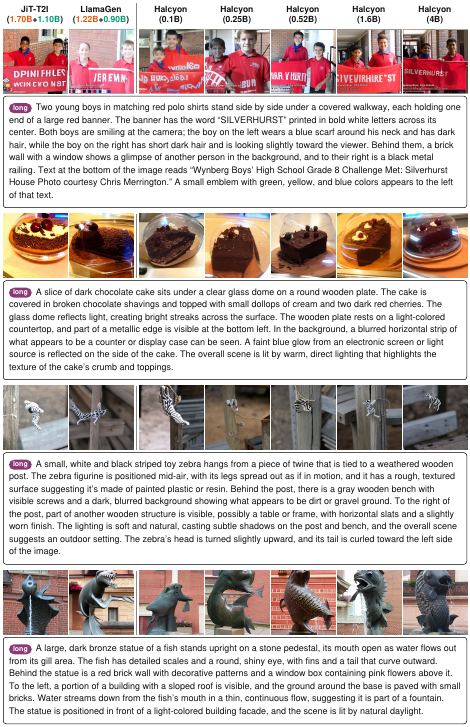}
    \vspace{-20pt}
    \caption{\textbf{Comparison of generated images with ``long" captions.}
    From left to right: JiT-T2I-XXL/8++, LlamaGen-XL/8, and \methodname{} from 0.1B to 4B.
    For the baselines, we report the sizes of the {\color{orange}\bf language model} and {\color{darkgreen}\bf image generator} separately.
    Image quality improves as \methodname{} scales up.
    All prompts are from GPIC Test-50K.
    }
    \label{fig:supp:collage_long}
\end{figure}

We compare generated samples across the four GPIC caption types: tag (Figure~\ref{fig:supp:collage_tag}), short (Figure~\ref{fig:supp:collage_short}), medium (Figure~\ref{fig:supp:collage_medium}), and long (Figure~\ref{fig:supp:collage_long}). Each row shows JiT-T2I-XXL/8++, LlamaGen-XL/8, and \methodname{}~from 0.1B to 4B for the same caption, sampled with the settings in Appendix~\ref{app:sampling}. Sample quality and prompt adherence improve as \methodname{}~scales, with larger models capturing finer details such as clothing, object attributes, and scene layout. Rendering legible text remains challenging for all models at 256×256 resolution.

